\documentclass[hf]{ceurart}

\usepackage[utf8]{inputenc} 
\usepackage[T1]{fontenc}    
\usepackage{hyperref}       
\usepackage{url}            
\usepackage{booktabs}       
\usepackage{amsfonts}       
\usepackage{nicefrac}       
\usepackage{microtype}      

\usepackage{amsmath,amsfonts,bm}

\def\eqref#1{equation~\ref{#1}}

\def\1{\bm{1}}

\def\va{{\bm{a}}}

\def\vs{{\bm{s}}}

\def\mO{{\bm{O}}}

\def\mW{{\bm{W}}}

\DeclareMathAlphabet{\mathsfit}{\encodingdefault}{\sfdefault}{m}{sl}
\SetMathAlphabet{\mathsfit}{bold}{\encodingdefault}{\sfdefault}{bx}{n}

\def\gG{{\mathcal{G}}}
\def\gH{{\mathcal{H}}}

\def\gL{{\mathcal{L}}}

\def\sH{{\mathbb{H}}}

\newcommand{\R}{\mathbb{R}}

\usepackage{graphicx}
\usepackage{amsmath}
\usepackage{amssymb}
\usepackage{subcaption}
\usepackage{tabularx}
\usepackage{listings}
\makeatletter
\def\fnum@lstlisting{%
  {\bfseries \lstlistingname
  \ifx\lst@@caption\@empty\else~\thelstlisting\fi}}%
\makeatother
\usepackage[capitalise]{cleveref}

\usepackage{biblatex}
\usepackage[firstpage, nostamp]{draftwatermark}
\SetWatermarkColor[gray]{0.9}

\begin{document}

\copyrightyear{2021}
\copyrightclause{Copyright for this paper by its authors.
  Use permitted under Creative Commons License Attribution 4.0
  International (CC BY 4.0).}

\conference{15th International Workshop on Neural-Symbolic Learning and Reasoning (NeSy)}

\title{pix2rule: End-to-end Neuro-symbolic Rule Learning}

\author[1]{Nuri Cingillioglu}[%
  email=nuric@imperial.ac.uk,
]
\author[1]{Alessandra Russo}[%
  email=a.russo@imperial.ac.uk,
]
\address[1]{Imperial College London, United Kingdom}

\begin{abstract}
  Humans have the ability to seamlessly combine low-level visual input with high-level symbolic reasoning often in the form of recognising objects, learning relations between them and applying rules. Neuro-symbolic systems aim to bring a unifying approach to connectionist and logic-based principles for visual processing and abstract reasoning respectively. This paper presents a complete neuro-symbolic method for processing images into objects, learning relations and logical rules in an end-to-end fashion. The main contribution is a differentiable layer in a deep learning architecture from which symbolic relations and rules can be extracted by pruning and thresholding. We evaluate our model using two datasets: subgraph isomorphism task for symbolic rule learning and an image classification domain with compound relations for learning objects, relations and rules. We demonstrate that our model scales beyond state-of-the-art symbolic learners and outperforms deep relational neural network architectures.
\end{abstract}


\maketitle

\vspace{-2em}
\section{Introduction}
Despite being surrounded by continuous input such as vision and sound, humans have evolved to recognise, process and maintain symbolic thought that seem to have co-evolved with the use of a symbolic natural language~\cite{coevoflangandbrain}. The result is a coherent information processing system which can integrate low-level signals with high-level abstract reasoning so well that symbolic cognitive models~\cite{cogmodelsymbolic} suggest the human mind operates on formal symbols. Similarly, physical symbol systems~\cite{physicalsymbolsystem} characterise cognition as not only symbol recognition but also manipulation and combination.
This symbolic thought on top of signal processing within a connectionist architecture is learnt at a young age with children building an understanding of objects, their relations and rules in their environment~\cite{piagetsymthought}. Furthermore, neurobiological mechanisms in the human brain have been proposed for handling triplets of symbols, e.g. in the form of subject, verb and object~\cite{brainaiqa}. Yet, this level of harmony between neural and symbolic domains remains a mystery for machine learning.

There has been a lot of interest and work on this topic under the umbrella of neuro-symbolic systems~\cite{neurosymsurvey}. In this context, neural refers to connectionist based approaches, mainly neural networks which have gained impressive achievements in image classification and audio recognition~\cite{deeplearningbook} that scale well with large amounts of data and computing power. The symbolic part often refers to a logical formalism such as first-order logic~\cite{russell2016artificial} and its implementation within a logic programming paradigm~\cite{logicprogramming}. Although in the early days of Artificial Intelligence symbolic approaches were prominent, particularly in the form of expert systems~\cite{expertsystems}, more recently they are used on top of neural networks in order to process non-symbolic sources of input such as images~\cite{nsldancunnington}. Our approach provides a new perspective on existing feed-forward neural networks to build an end-to-end architecture that can not only handle low-level input but also converge to symbolic relations and rules.

Hence, in this paper we focus on the question of whether an end-to-end neural network can learn objects, relations and rule-based reasoning going from pixels in an image all the way to logic-based rules.
Inspired by the perceptron algorithm~\cite{perceptron,russell2016artificial}, we define an inductive bias to an otherwise regular feed-forward neural network in order to enable end-to-end neuro-symbolic learning. The main contribution of this paper is a novel way of constraining the bias of a linear layer to obtain the semantics of AND and OR gates as well as pruning and thresholding to extract symbolic rules from that layer. We evaluate our approach in a controlled environment with two synthetic datasets and present the analysis of the learnt objects, relations and rules. Our implementation using TensorFlow~\cite{tensorflow} is publicly available at
  {\footnotesize \url{https://github.com/nuric/pix2rule}} with the accompanying data and analysis.

\vspace{-1em}
\section{Semi-symbolic Layer}\label{sec:sl_layer}
Our neuro-symbolic architecture makes use of a differentiable feed-forward layer that behaves like conjunction or disjunction.
Given continuous inputs $x_1, x_2, \ldots, x_n \in [\bot, \top]$ in which $\bot, \top \in \R$ denote some real-valued constants for false and true respectively, we would like to model a layer that can act like conjunction $y = \bigwedge_i x_i$ or disjunction $y = \bigvee_i x_i$ where $y \in [\bot, \top]$ is the output. If we take $\bot = 0$, $\top = 1$ and utilise t-norms~\cite{tnorms} to implement fuzzy logic, as the number of inputs $n$ increases, the operation suffers from vanishing gradients and becomes unviable for upstream layers such as convolutional neural networks (CNN)~\cite{deeplearningbook}. This phenomenon occurs due to the 1 out of $n$ failure or success characteristic of conjunction and disjunction respectively. Hence, we are interested in an operation that does not starve gradients and \emph{eventually} converges to the desired semantics.

Based on the single-layer perceptron~\cite{russell2016artificial}, we propose the semi-symbolic layer (SL):

\noindent\begin{minipage}{.5\linewidth}
  \begin{equation}\label{eq:slforward}
    y = f( \sum_i w_i x_i  + \beta )
  \end{equation}
\end{minipage}%
\begin{minipage}{.5\linewidth}
  \begin{equation}\label{eq:slbias}
    \beta = \delta ( \max_i |w_i| - \sum_i |w_i| )
  \end{equation}
\end{minipage}

where $w_i$ are the learnable layer weights, $f$ the non-linear activation function and $\delta \in [-1, 1]$ the semantic gate selector. In this formulation we set $\bot = -1$, $\top = 1$ and $f$ to be the hyperbolic tangent function (tanh). While \cref{eq:slforward} is the standard feed-forward layer, by adjusting the bias $\beta$ we can obtain either conjunctive when $\delta = 1$ or disjunctive when $\delta = -1$ semantics. Intuitively, in the conjunctive case, we are looking for a threshold that is at least as small as the sum of the weights but not any bigger than the maximum weight such that if at least one input is false, the output will be false too. For a more detailed derivation and an example implementation, please refer to \cref{apx:slderivation}.
By gradually adjusting $\delta$ from 0 to either 1 or -1, we can now maintain a feed-forward layer that does not starve gradients to upstream layers but eventually converges to the configured semantics. The sign of the weights indicate whether the input or its negation is contributing to the output. Thus, logical negation can be regarded as the multiplicative inverse of the input $\neg x_i = -x_i$ which naturally works with the weights of the layer. When there are no input connections, i.e. $\forall_i w_i = 0$ the output becomes zero and when there is only one input the bias becomes $\beta = 0$ yielding a pass-through, identity gate.

For symbolic inputs $x_i \in \{\bot, U, \top\}$ where $U$ is the symbol for unknown, $\delta \in \{1, -1\}$ and sufficiently large weights (large enough to saturate $\tanh$), the output of the semi-symbolic layer aligns with Łukasiewicz logic~\cite{manyvaluedlogics}. For example, in the conjunctive case when $\forall_i x_i = 0$ the output will be negative, \cref{eq:slbias}, implying that the conjunction of unknown inputs is false. In practice, when the input is a vector of learnt features from an upstream layer, the model has the chance to resolve unknown inputs if the task requires it. We leave further theoretic analyses of this layer's implied logics as future work.

\textbf{Prune \& Threshold} In order to obtain symbolic formulas, we prune and then threshold the weights after training. Similar to decision tree pruning methods~\cite{decisiontreepruning}, each weight is set to zero $w_i = 0$ and pruned if the performance has not dropped by a fixed $\epsilon$. Then a threshold value is picked by sweeping over potential values $[\min |w_i|, \max |w_i|]$ with a similar performance check to pruning. Each weight is then set to $w'_i = 6\operatorname{sign}(w_i)$ which gives sufficient saturation $\operatorname{tanh}(6) \approx 0.999$. Finally, we repeat the pruning step to remove any wrongly amplified weights. We opt for this simple post-training regime rather than regularisation such as l-norms~\cite{deeplearningbook} to limit the number of modifications required to train the semi-symbolic layer.

\vspace{-1em}
\section{Datasets}\label{sec:datasets}
\begin{figure}[t]
  \centering
  \begin{subfigure}[b]{0.47\textwidth}
    \centering
    \includegraphics[width=1.0\textwidth]{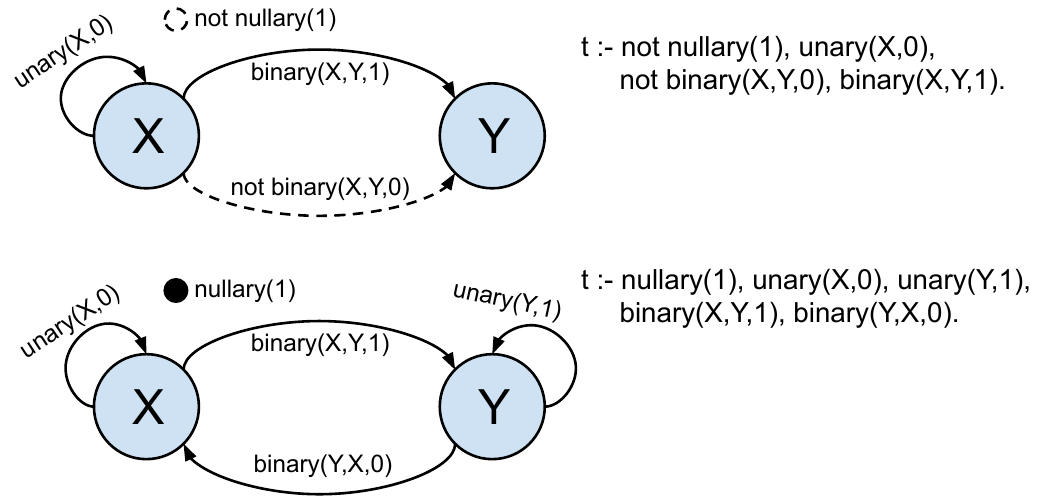}
    \caption{Sample target rules for the easy difficulty of the graph dataset. For each predicate, the last argument denotes the relation id.}
    \label{fig:gendnf_easy}
  \end{subfigure}
  ~
  \begin{subfigure}[b]{0.51\textwidth}
    \centering
    \includegraphics[width=0.9\textwidth]{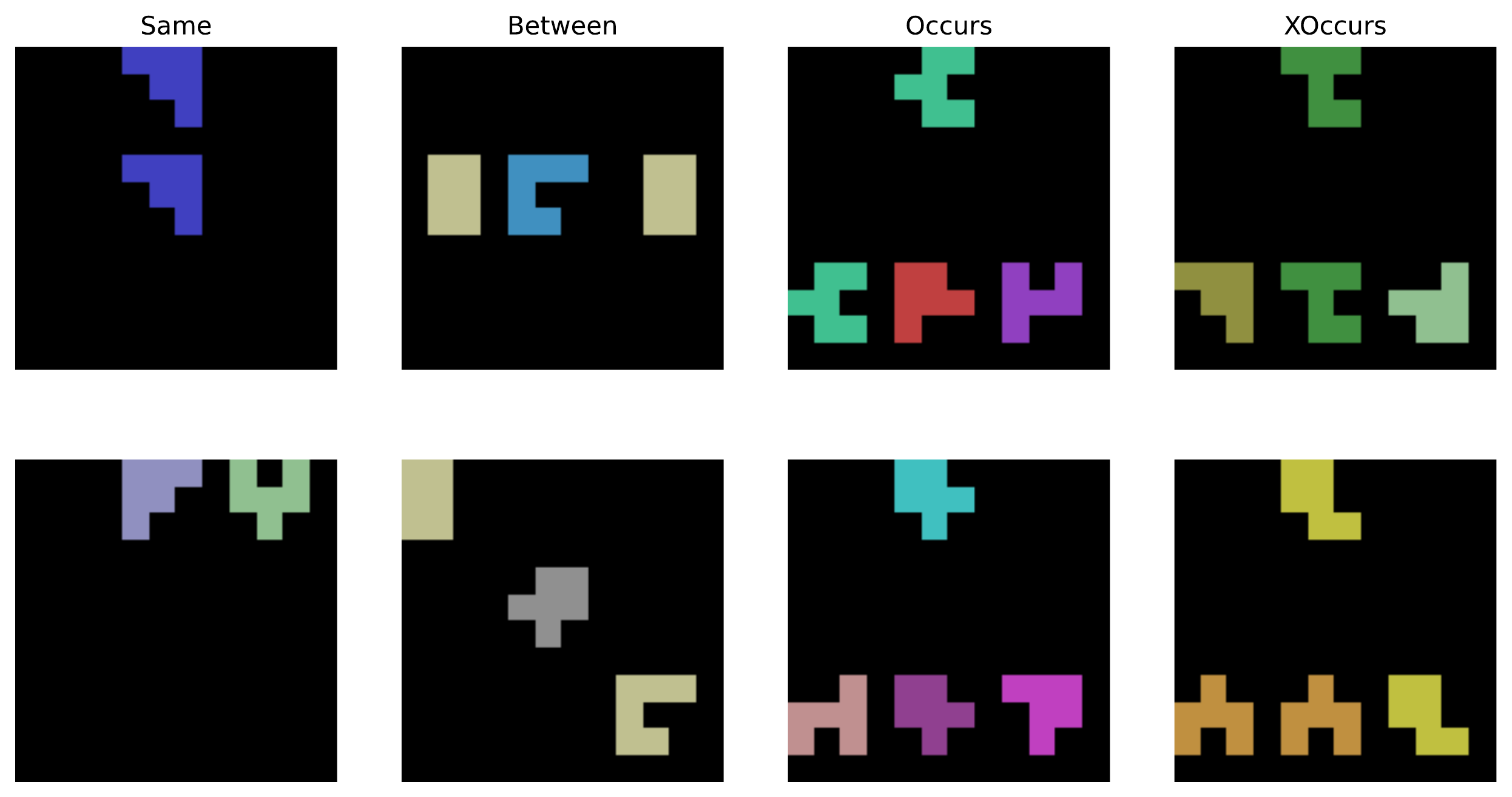}
    \caption{Example images from the relations game dataset with the top and bottom rows showing correct and incorrect cases respectively.}
    \label{fig:relsgame_tasks}
  \end{subfigure}
  \caption{Samples from datasets. For further details and examples please refer to \cref{apx:dataset_details}.}
  \vspace{-2em}
  \label{fig:datasets}
\end{figure}

Since low-level signals such as images can produce large noisy latent spaces, we need to understand whether our approach can robustly learn symbolic rules in a scalable fashion. Hence, we use two synthetic datasets: subgraph set isomorphism for testing the semi-symbolic layer over symbolic inputs and image classification for the full pipeline. For both datasets, the model is required to predict a true or false label. Samples from each dataset are shown in \cref{fig:datasets} with further details and examples in \cref{apx:dataset_details}. We opt to use synthetic datasets with known and controlled parameters to avoid any inherent biases that may arise in real-world datasets.

\textbf{Subgraph set isomorphism} requires a model to decide whether any subgraph of a given graph is isomorphic to a set of other graphs.
We choose graph isomorphism as it is combinatorial in nature and known to be NP-complete~\cite{npcomplete} providing a means to gauge scalability with respect to the search space of possible solutions.
Formally, given a graph $\gG$ and a set of graphs $\sH = \{ \gH_1, \gH_2, \ldots, \gH_n \}$, determine whether the target condition $t \Longleftrightarrow \exists_{\gL, i} \gL \subseteq \gG \land \gL \simeq \gH_i$ holds where $\simeq$ is graph isomorphism. Given graphs that either satisfy or do not satisfy $t$, the objective is to learn a $\sH$ for a fixed $n$. This learning task can be rendered as an instance of Inductive Logic Programming (ILP)~\cite{ilp} where $\sH$ is the hypothesis to learn with no background knowledge. We encode the problem in Answer Set Programming (ASP)~\cite{asp} using the generic predicates nullary/1 for global graph properties, unary/2 and binary/3 for self-edges and directed edges between nodes respectively with target condition $t$ as head of the rule, see \cref{fig:gendnf_easy}. Since we are interested in learning $\sH$ rather than just subgraph isomorphism, recent advancements in deep graph neural networks~\cite{graphnetworks, graphmatchingnetworks} are not suitable for extracting \emph{what} $\sH$ has been learnt, also known as the black-box problem~\cite{xaisurvey}.
To adjust the difficulty, \cref{tab:gendnf_difficulties}, we change the number of nodes $|V(\gG)|$, nullary, unary, binary relations, number of nodes in any $\gH_i$ and the maximum size of $|\sH|$ ensuring $|V(\gH_i)| \leq |V(\gG)|$. We focus on these parameters to alter the number of edges to be learnt $|E(\gH_i)|$ which in return corresponds to the length of the rules.
The average rule length in the medium difficulty is well beyond common datasets such as odd or even, family tree and graph colouring used in existing neuro-symbolic research~\cite{deltailp,logictensor,neurallogicnetworksilp}.

\begin{table}[h]
  \centering
  \caption{Different difficulty parameters for the subgraph set isomorphism dataset.}
  \label{tab:gendnf_difficulties}
  \begin{tabular}{@{}rccccccc@{}}
    \toprule
    Difficulty & $|V(\gG)|$ & Nullary & Unary & Binary & $|V(\gH_i)|$ & Max. $|\sH|$ & Avg. $|E(\gH_i)|$ \\
    \midrule
    Easy       & 3          & 2       & 2     & 2      & 2            & 3            & 7.29              \\
    Medium     & 4          & 4       & 5     & 6      & 3            & 4            & 37.27             \\
    Hard       & 4          & 6       & 7     & 8      & 3            & 5            & 50.62             \\
    \bottomrule
  \end{tabular}
\end{table}

\textbf{Relations Game} dataset consists of an input image with different shapes and colours exhibiting compound relations between them. It has been introduced to evaluate object-based relational learning in deep neural networks~\cite{predinet}. Existing methods fail to provide a coherent object, relation and symbolic rule learning in an end-to-end fashion. Along with the four tasks in \cref{fig:relsgame_tasks}, we create an All multi-task setting and provide the task id as additional input. While the training set contains pentominoes, shapes with 5 uniformly coloured pixels, the test sets expand to hexominoes (shapes with 6 pixels) as well as striped shapes with unseen colours. To ascertain if neuro-symbolic approaches could be more data efficient, we take 100, 1k and 5k for training and 1k for validation and test splits. We also apply standard data augmentation during training with random horizontal or vertical flips with an added input noise drawn from $\mathcal{N}(0, 0.01)$. For further details and more examples, please refer to \cref{apx:relsgame_details}.

\begin{figure}[t]
  \centering
  \includegraphics[width=1.0\textwidth]{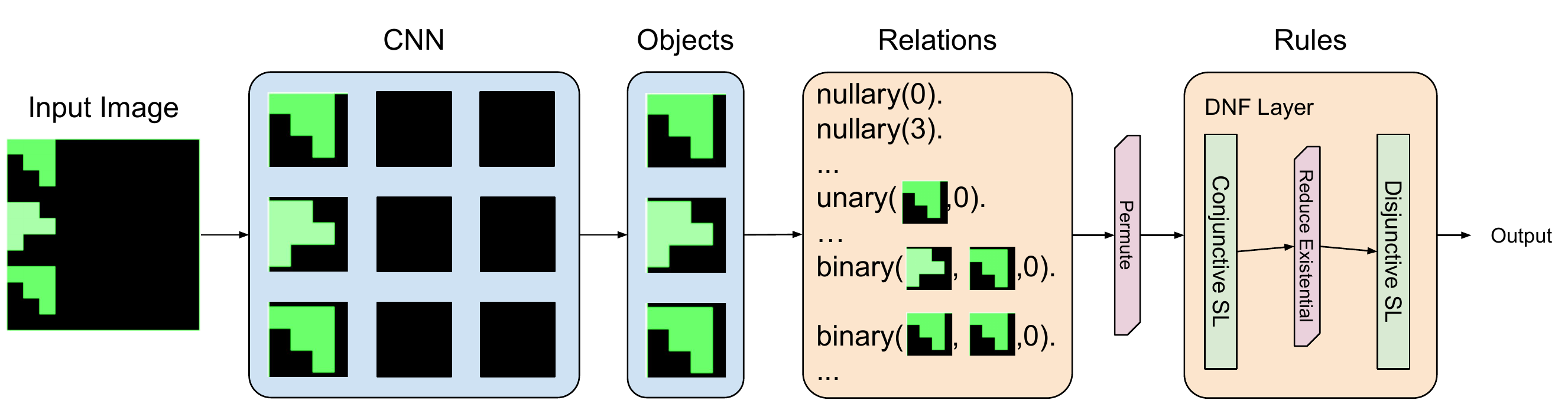}
  \caption{Graphical overview of the full neuro-symbolic model used for the relations game dataset. The image is processed as a 3x3 grid using a CNN to obtain representations for each patch. Then a subset of those patches are recognised as relevant objects using an MLP, see~\cref{apx:object_selection}. All relations between objects are computed using another MLP with shared weights. Every permutation is constructed to handle variable binding before passed on to the DNF layer to learn rules and predict the final output. Blue squares indicate distributional representations while the orange squares highlight fuzzy logic.}
  \vspace{-2em}
  \label{fig:relsgame_model}
\end{figure}

\vspace{-1em}
\section{Experiments}\label{sec:experiments}

We evaluate our approach's ability to learn symbolic rules in a differentiable manner, scalability and performance in the presence of input noise. To learn full logic rules such as in \cref{fig:gendnf_easy}, we stack two semi-symbolic layers, one conjunctive and one disjunctive, for a disjunctive normal form (DNF) layer. While theoretically different combinations of SL layers can extend the space of possible logic programs to a larger language including constraints and predicates with arity greater than two, we focus on learning first-order (recursive) normal logic programs without function symbols and excluding constraints. We also work with the nullary/1, unary/2 and binary/3 predicates since any logic program can be converted to a program using this signature.
The DNF layer is constructed with a maximum number of variables $|V(\gH_i)|$ and fixed $|\sH|$ to learn first-order rules rather than just propositional formula. When viewed as a fully-connected graph with variable symbols as the nodes (similar to \cref{fig:gendnf_easy}), the maximum number of edges $|E(\gH_i)|$ corresponds to the maximum length (number of body atoms) of the learnable rules. We gather all combinations $\binom{|V(\gG)|}{|V(\gH_i)|}$ to effectively curate every possible object to variable binding, shown as Permute step in \cref{fig:relsgame_model}. This grounding step converts first-order normal rules in question into their propositional instantiations over which SL layers from \cref{sec:sl_layer} are used. Although this binding is combinatorial, the evaluation of the rule under every binding can be computed in parallel which benefits from the increasing computing infrastructure available.
In cases where the desired rules require specific constants as literal arguments, the instantiations could be restricted at the permutation step to adjust the hypothesis space accordingly.
After the conjunctive SL, we use the $\max$ operator over the instantiations of any existential variables that may occur in the rules.
The DNF layer is used as a standalone model for the subgraph set isomorphism dataset or in tandem with upstream image processing layers for the relations game as in \cref{fig:relsgame_model}.
We train all deep models using Adam~\cite{adam} with a fixed learning rate of 0.001 and the negative log-likelihood loss on an Intel Core i7 CPU and report the median results to avoid outliers. For all the hyper-parameters and further training details, please refer to \cref{apx:hyperparameters}.

\begin{table}[h]
  \centering
  \caption{Median test accuracy for subgraph set isomorphism dataset without input noise with median absolute deviation. Although ILASP and FastLAS solve the easy set, they timeout at higher difficulties.}
  \label{tab:gendnf_ilp_vs_deep}
  \footnotesize
  \begin{tabular}{lrrrrrr}
    \toprule
    {}         & \multicolumn{3}{l}{Test Accuracy} & \multicolumn{3}{l}{Training Time}                                                                            \\
    Difficulty & Easy                              & Medium                            & Hard          & Easy               & Medium           & Hard             \\
    \midrule
    DNF        & 1.0$\pm$0.0                       & 1.00$\pm$0.0                      & 1.00$\pm$0.00 & 127.13$\pm$  4.17  & 136.90$\pm$10.00 & 129.56$\pm$ 5.77 \\
    DNF+t      & 1.0$\pm$0.0                       & 0.99$\pm$0.0                      & 0.99$\pm$0.01 & 125.02$\pm$  6.53  & 135.67$\pm$ 8.56 & 143.67$\pm$22.60 \\
    FastLASv3  & 1.0$\pm$0.0                       &                                   &               & 29.85$\pm$  0.69   &                  &                  \\
    ILASP-2i   & 1.0$\pm$0.0                       &                                   &               & 3336.00$\pm$994.45 &                  &                  \\
    \bottomrule
  \end{tabular}
\end{table}

\textbf{Can the DNF layer learn symbolic rules in a scalable, differentiable manner?} We train a standalone DNF layer on the subgraph set isomorphism dataset for 10k batch updates with a batch size of 128. The results with continuous weights (DNF) and with pruning, thresholding (DNF+t) are compared against two state-of-the-art symbolic learners: ILASP~\cite{ilasp} (2i as the recommended version for non-noisy tasks) and the recent more scalable FastLAS~\cite{fastlas}.
We also considered state-of-the-art rule mining system AMIE~\cite{amierulemining} but it does not support negated rules and uses breath-first search that does not scale well to long rules.
We were only able to run the symbolic learners for the easy size because they do not report progress or allow checkpointing making them infeasible for distributed shared computing infrastructure. FastLAS on an isolated machine did not terminate after 16 hours on the medium difficulty. In \cref{tab:gendnf_ilp_vs_deep}, we observe that DNF layer scales better than symbolic learners to larger rule sizes and since it is trained for fixed number of iterations maintains a steady training time. Since DNF+t is thresholded, we have exact logic semantics which correspond to the symbolic rules learnt by the symbolic systems. We can convert the thresholded weights into ASP rules and test them using clingo~\cite{clingo5} to verify that our approach has indeed learnt correct symbolic rules in a differentiable manner.

\begin{table}[ht]
  \centering
  \caption{Median test accuracy for the best out of 5 runs with input noise. The median absolute deviation is less than 0.09 for all entries.}
  \label{tab:gendnf_results}
  \footnotesize
  \begin{tabular}{lrrrrrrrrr}
    \toprule
    Difficulty & \multicolumn{3}{l}{Easy} & \multicolumn{3}{l}{Medium} & \multicolumn{3}{l}{Hard}                                           \\
    Noise      & 0.00                     & 0.15                       & 0.30                     & 0.00 & 0.15 & 0.30 & 0.00 & 0.15 & 0.30 \\
    \midrule
    DNF        & 1.00                     & 0.89                       & 0.82                     & 1.00 & 0.98 & 0.89 & 1.00 & 0.98 & 0.86 \\
    DNF+t      & 1.00                     & 1.00                       & 0.84                     & 0.99 & 0.99 & 0.99 & 0.99 & 0.99 & 0.74 \\
    \bottomrule
  \end{tabular}
\end{table}

\textbf{How does the DNF layer cope with input noise?} We perturb the input since learnt latent representations of low-level signals are likely to be noisy. Hence, we add input noise to the subgraph set isomorphism dataset by randomly flipping the truth values of $E(\gG)$ in the training set with a fixed probability shown in \cref{tab:gendnf_results}. We observe that the DNF layer performs well up to 0.3 where the median accuracy drops below 0.9. The pruning and thresholding steps seem to improve the performance with lower levels of noise, likely because incorrect weights are removed against a non-noisy validation set.

\begin{table}[h]
  \centering
  \caption{Median test accuracy for relations game tasks. For full results, please refer to \cref{apx:further_results}.}
  \label{tab:relsgame_results}
  \tiny
  \begin{tabular}{llrrrrrrrrrrrrrrr}
    \toprule
           & Task     & \multicolumn{3}{l}{All} & \multicolumn{3}{l}{Between} & \multicolumn{3}{l}{Occurs} & \multicolumn{3}{l}{Same} & \multicolumn{3}{l}{XOccurs}                                                                       \\
    Set    & Model    & 100                     & 1000                        & 5000                       & 100                      & 1000                        & 5000 & 100  & 1000 & 5000 & 100  & 1000 & 5000 & 100  & 1000 & 5000 \\
    \midrule
    Hex.   & DNF      & 0.94                    & 0.97                        & 0.98                       & 0.90                     & 0.99                        & 0.99 & 0.56 & 0.99 & 0.99 & 0.94 & 1.00 & 1.00 & 0.49 & 0.80 & 0.93 \\
           & DNF-h    & 0.98                    & 0.99                        & 0.99                       & 0.95                     & 1.00                        & 0.99 & 0.62 & 0.99 & 0.99 & 0.97 & 1.00 & 1.00 & 0.50 & 0.98 & 0.96 \\
           & DNF-h+t  & 0.51                    & 0.55                        & 0.92                       & 0.91                     & 0.97                        & 0.98 & 0.50 & 0.79 & 0.96 & 0.53 & 1.00 & 0.98 & 0.48 & 0.51 & 0.51 \\
           & DNF-hi   & 0.98                    & 0.99                        & 1.00                       & 0.97                     & 0.99                        & 1.00 & 0.69 & 0.99 & 0.99 & 0.97 & 1.00 & 1.00 & 0.51 & 0.99 & 0.99 \\
           & DNF-r    & 0.94                    & 0.98                        & 0.98                       & 0.84                     & 1.00                        & 0.99 & 0.63 & 0.99 & 0.99 & 0.96 & 1.00 & 1.00 & 0.51 & 0.94 & 0.51 \\
           & PrediNet & 0.85                    & 0.95                        & 0.96                       & 0.66                     & 0.99                        & 0.99 & 0.57 & 0.95 & 0.97 & 0.99 & 1.00 & 1.00 & 0.50 & 0.58 & 0.95 \\
    Pent.  & DNF      & 0.89                    & 0.96                        & 0.95                       & 0.85                     & 0.99                        & 0.99 & 0.57 & 0.96 & 0.98 & 0.95 & 1.00 & 1.00 & 0.50 & 0.74 & 0.86 \\
           & DNF-h    & 0.95                    & 0.97                        & 0.98                       & 0.92                     & 0.99                        & 0.99 & 0.62 & 0.95 & 0.97 & 0.94 & 1.00 & 1.00 & 0.51 & 0.94 & 0.90 \\
           & DNF-h+t  & 0.50                    & 0.53                        & 0.88                       & 0.90                     & 0.98                        & 0.97 & 0.49 & 0.81 & 0.93 & 0.50 & 0.99 & 0.97 & 0.50 & 0.51 & 0.49 \\
           & DNF-hi   & 0.96                    & 0.99                        & 0.99                       & 0.95                     & 0.99                        & 1.00 & 0.69 & 0.98 & 0.98 & 0.96 & 1.00 & 1.00 & 0.50 & 0.96 & 0.99 \\
           & DNF-r    & 0.93                    & 0.96                        & 0.97                       & 0.81                     & 0.99                        & 0.99 & 0.65 & 0.96 & 0.96 & 0.95 & 1.00 & 1.00 & 0.52 & 0.88 & 0.49 \\
           & PrediNet & 0.85                    & 0.96                        & 0.95                       & 0.65                     & 0.99                        & 0.98 & 0.60 & 0.95 & 0.97 & 0.99 & 1.00 & 1.00 & 0.50 & 0.58 & 0.95 \\
    Stripe & DNF      & 0.91                    & 0.97                        & 0.95                       & 0.81                     & 0.98                        & 0.99 & 0.57 & 0.97 & 0.99 & 0.93 & 0.99 & 1.00 & 0.49 & 0.88 & 0.94 \\
           & DNF-h    & 0.93                    & 0.98                        & 0.99                       & 0.89                     & 0.99                        & 0.99 & 0.57 & 0.97 & 0.99 & 0.96 & 1.00 & 1.00 & 0.51 & 0.98 & 0.97 \\
           & DNF-h+t  & 0.52                    & 0.53                        & 0.93                       & 0.92                     & 0.97                        & 0.95 & 0.49 & 0.84 & 0.86 & 0.49 & 0.99 & 0.97 & 0.48 & 0.50 & 0.51 \\
           & DNF-hi   & 0.95                    & 0.99                        & 0.99                       & 0.94                     & 0.99                        & 0.99 & 0.63 & 0.96 & 0.99 & 0.97 & 1.00 & 1.00 & 0.49 & 0.96 & 0.97 \\
           & DNF-r    & 0.92                    & 0.97                        & 0.98                       & 0.81                     & 0.99                        & 0.99 & 0.64 & 0.95 & 0.98 & 0.98 & 1.00 & 1.00 & 0.52 & 0.92 & 0.51 \\
           & PrediNet & 0.84                    & 0.93                        & 0.92                       & 0.64                     & 0.99                        & 0.99 & 0.54 & 0.94 & 0.92 & 0.99 & 0.99 & 1.00 & 0.51 & 0.61 & 0.93 \\
    \bottomrule
  \end{tabular}
\end{table}

To perform neuro-symbolic reasoning with images, we now combine the DNF layer with upstream convolutional neural networks (CNN) and object selection layers to construct the full neuro-symbolic DNF model (DNF), \cref{fig:relsgame_model}. The image is processed in patches by the CNN to obtain a set of 9 candidate objects. Since reasoning with 9 objects at the same time creates a large grounding for the DNF layer, we use an attention based object selection layer.
Attention models~\cite{bahdanauatt, attsurvey} allow neural networks to focus on specific parts of the input, in this case selecting relevant patches of the given input image. The attention map is used as the parameters of a Gumbel-Softmax~\cite{gumbelsoftmax}, also known as Concrete~\cite{concretedistribution}, distribution to gradually sharpen the attention maps and have a clear correspondence between the selected objects and the rules applied thereafter.
We leave more complex, unsupervised scene object recognition and decomposition methods~\cite{monet,slotattention,graphscenedecomposition} for future work. We then use a single feed-forward layer with shared weights to compute all relations between the selected objects, and pass onto a DNF layer. For further details please refer to \cref{apx:model_details}. We construct 2 additional variants: DNF-h has extra hidden DNF layers, one for each 14 invented predicates, and DNF-r iterates the DNF layer twice learning recursive rules with 7 invented predicates. The invented predicates are evaluated in parallel using matrix operations similar to a feed-forward layer with many outputs.

\textbf{Is the DNF model more data efficient?} We compare our approach against PrediNet~\cite{predinet}, an explicitly relational neural network and the current state-of-the-art in the relations game dataset. Note that PrediNet already outperforms MLP baselines and Relation Networks~\cite{relationnetworks}. Looking at columns with different data sizes in \cref{tab:relsgame_results}, we observe that the DNF models consistently match or outperform PrediNet, especially DNF-h with training size 100 for the between task. This gap eventually narrows as the training size is increased.
Despite failing to recognise and reason with four objects in the Occurs and XOccurs tasks with 100 training examples, all the models seem to tackle them better when mixed with other tasks, i.e. the All task. This might suggest that a mixture of tasks is beneficial to learning such that tasks with 2, 3 and 4 objects are presented together.. The failure cases observed with 100 training data points highlight how our approach is still prone to over-fitting due to its neural network based formulation which may be further mitigated using standard techniques such as regularisation.

\textbf{Can the DNF model generalise to unseen shapes and colours?} After training on pentominoes, the models are tested on unseen hexominoes and striped shapes, \cref{apx:relsgame_details}. The results for Hex. and Stripe in \cref{tab:relsgame_results} are similar to that of pentominoes for all models. This lack of change might be because of the tasks' requirement to determine whether the shapes match or not, which can encourage a simple subtraction based representation that would generalise to unseen shapes and colours. Indeed, PrediNet uses subtraction as the main relational operator while our model would need to learn that solely from examples.

\textbf{Does image reconstruction improve DNF model performance?} We try to reconstruct the image from the selected objects using deconvolutional layers as an auxiliary loss (DNF-hi), see \cref{apx:image_reconstruction} for reconstruction details and \cref{apx:further_results} for examples. Comparing DNF-hi and DNF-h rows, we do not observe any improvement above 5\% despite the extra computation required to reconstruct the images. Although the representations of objects would need to accommodate both the reasoning and the reconstruction, we believe the simplicity of the shapes might be why the auxiliary loss does not yield any advantage for the rule learning.

\textbf{Can we extract symbolic rules in an image classification task?} Finally, the goal is to obtain a coherent neuro-symbolic pipeline where objects, their relations and symbolic rules are learnt. We take the best variant DNF-h and threshold its weights DNF-h+t to get symbolic rules. Comparing DNF-h with DNF-h+t in \cref{tab:relsgame_results}, only if DNF-h achieves $\geq 0.99$ accuracy does DNF-h+t consistently yield better than random performance.
This highlights the difficulty of learning both the predicates and the rules in tandem as well as the fragility of thresholding weights.
Yet, there are some successful outliers ignored by the median, shown in \cref{apx:further_results}.

\vspace{-1em}
\section{Analysis}\label{sec:analysis}

To understand what the model has learnt and how it is behaving, we take a single successful run ($\geq 0.95$ accuracy) from the DNF model with image reconstruction on the between task with 1k training examples. We choose this task because it is smaller in size and the model achieves 0.99 and 0.97 test hexominoes accuracy prior to and after thresholding respectively. For further examples and analysis, please refer to \cref{apx:further_results}.

\begin{figure}[t]
  \centering
  \begin{subfigure}[b]{0.45\textwidth}
    \centering
    \includegraphics[width=1.00\textwidth]{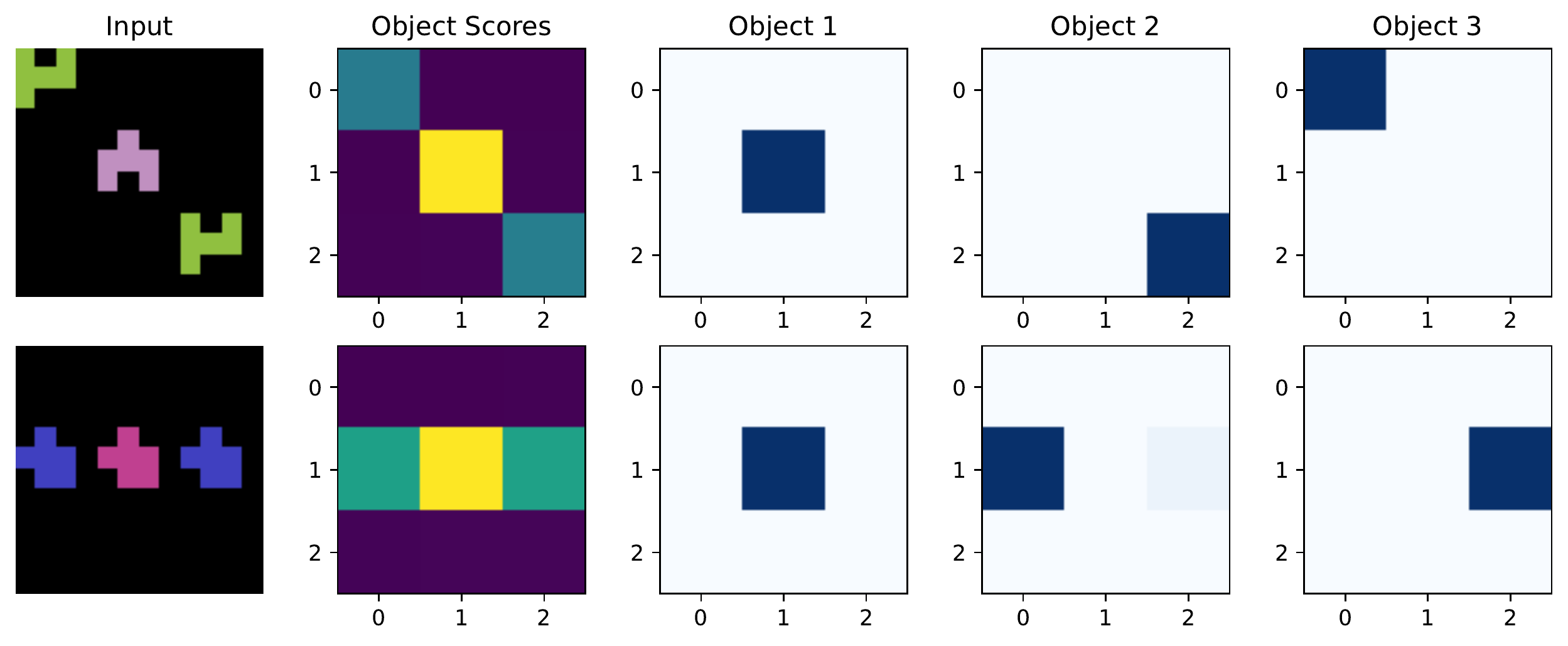}
    \caption{Attention maps learnt during the object selection phase. Gumbel-Softmax with low temperature approximates hard attention.}
    \label{fig:relsgame_DNF-i_between_att_maps}
  \end{subfigure}
  ~
  \begin{subfigure}[b]{0.53\textwidth}
    \centering
    \includegraphics[width=1.00\textwidth]{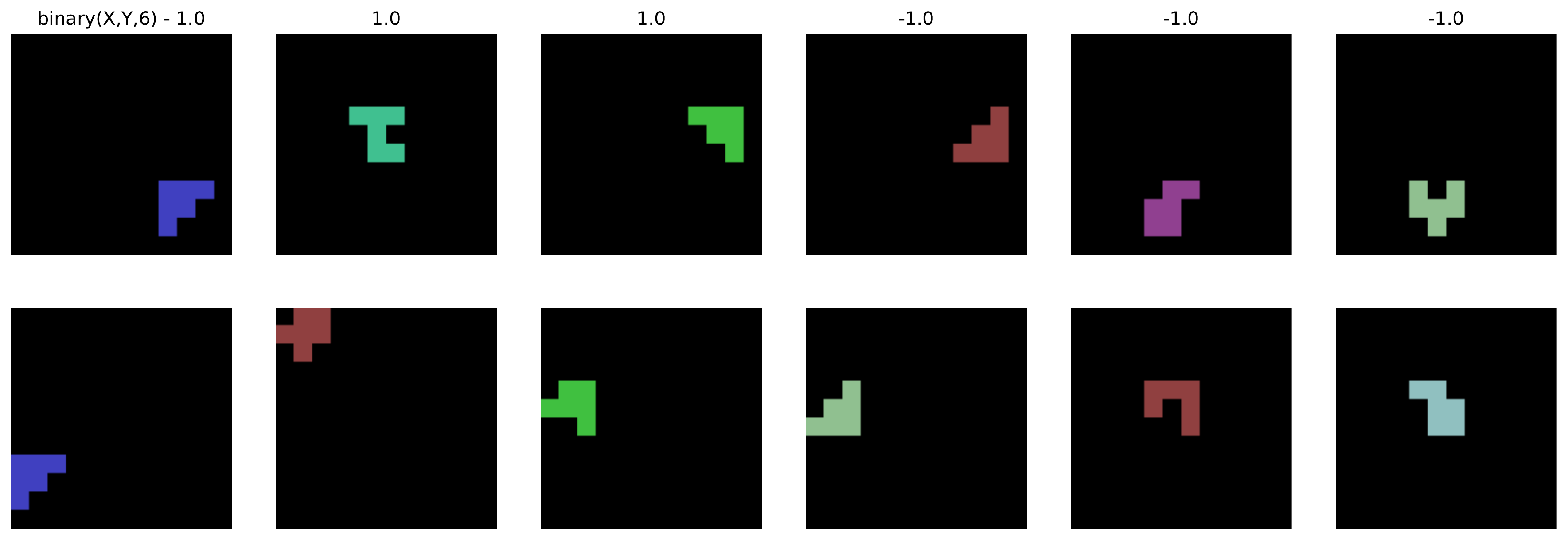}
    \caption{Object arguments (top and bottom rows) that make \texttt{binary(X,Y,6)} true or false exhibit no common pattern of the principal concepts of the dataset.}
    \label{fig:relsgame_binary6_truth_cases}
  \end{subfigure}
  \caption{Analysis plots of a single run of the DNF model trained on between task with 1k training examples and image reconstruction. Further analysis can be found in \cref{apx:further_results}.}
  \vspace{-2em}
  \label{fig:relsgame_analysis}
\end{figure}

\textbf{What are the selected objects?} To ensure whether the relations and rules actually work with the desired objects, we plot the attention maps obtained from the object selection step. As shown in \cref{fig:relsgame_DNF-i_between_att_maps}, the model learns to assign higher scores to patches with objects and iteratively selects them. Due to the random sampling, the order of the selected objects may vary.

\textbf{In a successful run, what are the rules learnt?} Since the thresholded model has a one-to-one correspondence with the desired logic semantics, we extract out the rule used to solve the between task, e.g. input images shown in \cref{fig:relsgame_DNF-i_between_att_maps}:

\begin{center}
  \begin{tabular}{|rl|}
    \toprule
    between :- & unary(X,4), not unary(Y,6), binary(Y,Z,6), not binary(Y,Z,9),        \\
               & not binary(Z,Y,3), binary(Z,Y,5), not binary(Z,Y,9), binary(Z,Y,14). \\
    \bottomrule
  \end{tabular}
\end{center}

which achieves 0.97 accuracy on the test hexominoes set. This is the first result we are aware of that presents differentiable rule learning with learnt predicates on continuous representations of images in an end-to-end fashion. To further validate this rule, we threshold $>0$ the interpretation, relations box in \cref{fig:relsgame_model}, and pass them, along with the rule to clingo. Over a sample test batch of 64 examples, clingo solves 91\% of them verifying the correctness of the learnt rules by the DNF layer.
This seamless integration of symbolic relations and rules with distributed representations of objects reflect on how the human brain might be using symbolic and sub-symbolic representations in tandem~\cite{symandsubsyminteg}.

\textbf{What do the learnt predicates mean?} We iteratively remove one atom from the body and check the drop in accuracy over the same sample test batch. Removing binary(X,Y,6) yields the highest drop of 18\%. Hence, we plot cases of binary(X,Y,6) in \cref{fig:relsgame_binary6_truth_cases} but fail to notice any alignment between the primary concepts of the dataset such as shape, colour or position. Although at first this may seem odd, since there is no regularisation for any sort of disentanglement, we would not expect or assume any correlation between the concepts and the learnt relations.
This is in contrast with explainable AI~\cite{xaisurvey} and the promise of interpretability using neuro-symbolic methods which is fundamentally limited by our ability to decode learnt representations.
On the other hand, this phenomenon does not arise when the input predicates are known and fixed such as in the graph isomorphism dataset where the learnt rules have interpretable meanings in the form of graphs, see~\cref{fig:gendnf_easy}.

\vspace{-1em}
\section{Related Work}\label{sec:related_work}

Neuro-symbolic~\cite{neurosymbolic} architectures and methods have a prolific history. We can categorise related work based on how neural or symbolic they are.
On the far neural end of the spectrum, we can consider deep neural networks designed to perform symbolic manipulation such as Neural Turing Machines~\cite{neuralturing} and its successor the Differentiable Neural Computer~\cite{dnc}. They attempt to create universal computers that can be trained and, in theory, perform symbol manipulation for example by solving algorithmic tasks. However, \emph{how} they perform symbolic representation and manipulation is not clear. Studies of algorithmic tasks using recurrent neural network based controllers suggest limitations in the extent of symbolic manipulation by continuous dense representations~\cite{learnsimplealgo}. This symbolic manipulation can also extend to program execution with Neural-program interpreters~\cite{neuralproginterpreters} and Neural Symbolic Machines~\cite{neuralsymbolicmachines} which operate in discrete steps but use reinforcement learning in doing so. Discrete data-structures such as stacks have also been used in aid of this problem such as Stack-augmented Recurrent Nets~\cite{neuralstack}.

Going one step closer to symbolic, architectural biases are often added to constrain and refine the behaviour of neural networks. For example, Relation Networks~\cite{relationnetworks} create a pairwise object representation bottleneck and demonstrate improvements in visual question answering tasks. This architectural inductive bias can be traced to many models all the way and including PrediNet~\cite{predinet}. More recently, an application of rule based systems that also use Gumbel-Softmax~\cite{gumbelsoftmax} to sharpen the rule selection was proposed with Neural Production Systems~\cite{neuralproductionsystems}. Unsupervised deep representation learning using auto-encoders were also shown to work in the planning domain where the latent representation was converted to propositional atoms~\cite{planningdeeplatent, planningdeeplatentextended}. In relation to the subgraph isomorphism problem, one can also deploy graph neural networks~\cite{graphnetworks} that utilise message passing between nodes. In particular, Graph Matching Networks~\cite{graphmatchingnetworks} are designed to determine graph similarities but it remains unknown how the matched graphs could be extracted from continuous weights.

Methods described so far do not consider logical formalisms or logic programs. DeepLogic~\cite{deeplogic} provides a first insight into attempts of learning symbolic reasoning in an end-to-end fashion using memory networks~\cite{memnn, memn2n} which has gated recurrent-units (GRU)~\cite{gru} as the basis of symbolic manipulation.
Neural Logic Machines~\cite{neurallogicmachines} use principles of forward-chaining~\cite{russell2016artificial} with nullary, unary and binary predicates implemented as MLPs with input permutations to tackle algorithmic tasks. However, due to the use of MLPs, the learnt rules or reasoning steps cannot be symbolically extracted.
Our approach falls in this section of the spectrum: constrain or make neural networks exhibit behaviours of logical reasoning. That is, to build an architecture with the right biases in order to learn and leverage structured logical reasoning in the form of objects, relations and rules.

Going further away from deep neural networks, logic based architectures become more prominent. For example, Lifted Relational Neural Networks~\cite{liftedneuralnetworks} use rule sets as templates for constructing neural networks, i.e. the connection paths. Logic Tensor Networks~\cite{logictensor} use dense embeddings of constants while also constructing a deductive neural network. TensorLog~\cite{tensorlog} similarly builds factor graphs which in return yield the neural network architecture.
These approaches have similarities to the more recent Logical Neural Networks~\cite{ibmlogicalneuralnetworks} which assemble neural networks from logical formulae and constrain the weights to achieve conjunction and disjunction semantics. Our more flexible approach is not bound to a fixed logic program and does not require any constraints on the weights whilst also handling negation.

One could also use distributed representations for constants or predicates. For example similarity between vectors could provide basis for a logical calculus~\cite{timlowdim}. Neural Theorem Provers~\cite{timntp} learn embeddings of predicates by unrolling given logic programs using backward-chaining~\cite{backwardchaining}, effectively following the steps of symbolic reasoning prior to any learning. These works have been inspired by word embeddings~\cite{word2vec, glove, wordembedssurvey} that capture distributional semantics of words. More recently, Neural Datalog~\cite{neuraldatalog} uses vector representations along with a datalog program to improve performance of temporal modelling.

Closer to the symbolic end of the spectrum, one could also attempt to decompose or parse continuous input completely into symbolic entities or concepts and then perform symbolic reasoning. Neural scene parsing as done in Neuro-Symbolic VQA~\cite{neurosymbolicvqa} and following works, completely decompose images into objects, their properties before performing visual question answering. For synthetic datasets where the decomposition is successful, it outperforms any end-to-end approach. We consider scene parsing closer to symbolic approaches since the reasoning about objects and their relations is not done by neural networks but by the manually engineered reinforcement learning environment with pre-defined functions. Neuro-symbolic Concept Learner~\cite{neurosymbolicconceptlearner} extends scene decomposition by learning concepts given in a parsed question such as red in a joint fashion using reinforcement learning in which the environment is a symbolic program executor.

Moving away from neural networks but attempting to harness gradient descent, we have approaches such as $\delta$-ILP~\cite{deltailp} and derivations~\cite{functionaldeltailp} which implement t-norms to create differentiable logic programs to find a suitable hypothesis in an ILP setting. More recently, by directly modelling rule membership of atoms as learnable weights, Neural Logic Networks~\cite{neurallogicnetworks, neurallogicnetworksilp} provide a competitive differentiable ILP system that leverages gradient descent.
The influence of gradient descent can also be found in full symbolic reasoning systems like DeepProblog~\cite{deepproblog} and NeurASP~\cite{neurasp} which attempt to propagate gradients through logic programs in order to train input neural networks such as CNNs that recognise hand-written digits and perform addition of the recognised digits.
This resembles continuous formulations of logic such as differentiable stable and supported semantics that utilise matrix multiplications and non-linear activations to realise logic operations~\cite{yanivvectorspaces}.

\vspace{-1em}
\section{Conclusion}

We presented a unified neuro-symbolic framework for learning objects, their relations and symbolic rules in an end-to-end fashion using semi-symbolic layers. Evaluation on two datasets portray competitive if not better results of our approach against symbolic learners and deep neural networks. Since symbolic rules can be extracted and examined, we plan to apply this technique to more decision making critical domains such as reinforcement learning.

\begin{acknowledgments}
  We would like to thank Murray Shanahan for his helpful comments. We would also like to thank Mark Law for his support in running ILASP and FastLAS symbolic learners.
\end{acknowledgments}

\AtNextBibliography{\small}
\printbibliography

\appendix

\section{Model Details}\label{apx:model_details}
We use two models in our experiments that share the common DNF layer described in \cref{sec:sl_layer} and \cref{sec:experiments}. In this section, we provide details on the semi-symbolic layer as well as the larger model used for the image classification task.

\subsection{Semi-symbolic Layer}\label{apx:slderivation}
The semi-symbolic layer builds on the semantics of how a regular single-layer perceptron~\cite{russell2016artificial} behaves in order to achieve AND semantics. Let's start with the formulation for the pre-activation value for a single-layer perceptron:

\begin{equation}
  \sum_i w_ix_i + \beta = z
\end{equation}

Now, suppose we are interested in obtaining AND gate semantics with a pre-activation value $z$ if all inputs are true and $-z$ if one of the inputs are false. To simplify the derivation, suppose the magnitudes of the weights are equal $\forall_{ij} w_i = w_j$:

\begin{align}
  \sum_i |w_i| + \beta         & = z  \\
  \sum_i |w_i| - |w_i| + \beta & = -z
\end{align}

where a true input means $x_i = 1.0$ if the weight is positive or $x_i = -1.0$ otherwise. We use the absolute value of the weights since the sum of the weights multiplied by matching inputs would yield the sum of the absolute value of the weights. Solving the above system of equations for $\beta$, we obtain:

\begin{align}
  2\beta & = |w_i| - 2\sum_i |w_i|          \\
  \beta  & = \frac{|w_i|}{2} - \sum_i |w_i|
\end{align}

Since the weights would not be equal during training, we replace the first term to obtain the final equation \cref{eq:slbias} presented in \cref{sec:sl_layer}:

\begin{equation}
  \beta = \max_i |w_i| - \sum_i |w_i|
\end{equation}

which preserves the desired AND gate semantics. The derivation for the disjunctive case is identical and leads to the bias terms flipped, i.e. sum - max. A fully functional implementation of the semi-symbolic layer in TensorFlow can be found in \cref{list:sl_source}.

\begin{lstlisting}[language=Python,caption={Implementation of semi-symbolic layer in Tensorflow.},label={list:sl_source},basicstyle=\footnotesize]
import tensorflow as tf

def semi_symbolic(in_tensor: tf.Tensor, kernel: tf.Tensor, delta: float):
    """Compute semi-symbolic layer outputs of a given input tensor."""
    # in_tensor (..., H), kernel (H,), delta [1,-1]
    abs_kernel = tf.math.abs(kernel) # (H,)
    bias = tf.reduce_max(abs_kernel) - tf.reduce_sum(abs_kernel) # ()
    conjuncts = tf.reduce_sum(in_tensor * kernel, -1) + delta*bias
    return tf.nn.tanh(conjuncts)
\end{lstlisting}

\subsection{Input Image CNN}\label{apx:input_image_cnn}

To process the input images in the relations game dataset, \cref{sec:datasets}, we use a single layer convolutional neural network (CNN) with a kernel size of 4x4, stride 4, relu activation and 32 filters. We use these options to keep the image processing layer simple and less computationally intensive. The input to the layer is a single relations game image, 12x12x3 and the output is 3x3x32. Note that, with this configuration the CNN is aligned to the grid cells of the image in which the objects are contained.

For each output of the CNN, 3x3x32, we also append the location information yielding a final result of 3x3x36. The coordinates are computed on a linear scale from 0 to 1 based on the location of the patch in the 3x3 grid. The representations are flattened into 9x36 and passed onto the selection layer. Example representations learnt by this layer are shown in \cref{tab:obj_reprensentation}.

\begin{table}
  \centering
  \caption{Example continuous representations of objects that are selected. The last 4 entries are the location coordinates appended based on the object's 3x3 grid location.}
  \label{tab:obj_reprensentation}
  \footnotesize
  \begin{tabular}{r|llllllllllll}
    \toprule
    Obj 1 & 0.    & 3.356 & 0.728 & 0.    & 0.871 & 1.170 & 1.726 & 1.528 & 1.174 & 0.117 & 1.586 & 0.478 \\
          & 2.527 & 0.286 & 0.724 & 1.723 & 0.580 & 0.    & 3.770 & 0.943 & 0.517 & 3.114 & 3.456 & 0.    \\
          & 0.710 & 0.975 & 2.030 & 1.570 & 1.251 & 1.971 & 1.830 & 2.688 & 0.5   & 0.5   & 0.5   & 0.5   \\
    \hline
    Obj 2 & 2.624 & 0.304 & 0.244 & 0.    & 0.    & 2.959 & 0.589 & 3.234 & 2.745 & 2.195 & 0.    & 2.967 \\
          & 0.510 & 2.248 & 2.414 & 1.953 & 2.047 & 1.409 & 1.645 & 2.542 & 0.    & 2.825 & 1.280 & 0.    \\
          & 2.709 & 2.844 & 0.293 & 0.349 & 3.549 & 0.    & 0.    & 0.    & 1.    & 1.    & 0.    & 0.    \\
    \hline
    Obj 3 & 2.624 & 0.304 & 0.244 & 0.    & 0.    & 2.959 & 0.589 & 3.234 & 2.745 & 2.195 & 0.    & 2.967 \\
          & 0.510 & 2.248 & 2.414 & 1.953 & 2.047 & 1.409 & 1.645 & 2.542 & 0.    & 2.825 & 1.280 & 0.    \\
          & 2.709 & 2.844 & 0.293 & 0.349 & 3.549 & 0.    & 0.    & 0.    & 0.    & 0.    & 1.    & 1.    \\
    \bottomrule
  \end{tabular}
\end{table}

\subsection{Object Selection}\label{apx:object_selection}
Given a set of objects with continuous representations in the form of a matrix $\mO \in \R^{n \times d}$ where $n$ is the number of objects and $d$ is the number of dimensions, the object selection layer selects $m$ many relevant objects. The number of objects $m$ is fixed and determined by the task. We use 2 for Same, 3 for Between and 4 for the rest of the tasks including All. The \emph{relevance}, or score, is learnt using single feed-forward layer. Concretely, the 9x36 object matrix obtained from the CNN layer, is mapped to a vector $\vs \in \R^9$. Then these unnormalised scores are used as logits for a Gumbel-Softmax~\cite{gumbelsoftmax} or Concrete~\cite{concretedistribution} distribution, from which we sample once to obtain an attention map $\va \in \R^m$. We then use an iterative \emph{score inversion} principle:
\begin{equation}
  \vs^{t+1} = \va(\vs - c) + (1-\va)\vs
\end{equation}
where $c=100$ is the inversion constant. We iterate for $m$ times selecting one object at each iteration and by inverting its score, the layer learns to select $m$ distinct objects.

Similar to previous work that use Gumbel-Softmax as a means of differentiable categorical distribution~\cite{planningdeeplatent}, we anneal the temperature of the distribution starting from 0.5, down to 0.01 with an exponential rate of 0.9 every step after 20 epochs. That allows the model to gradually learn and then sharpen the object selection process. At a temperature of 0.01, the attention maps become one-hot vectors allowing a clear correspondence between the selected objects and the applied rules. This is in contrast with soft attention~\cite{bahdanauatt} mechanisms such as the dot-product attention used in PrediNet which allow continuous selection of image patches as a single \emph{object}.

\subsection{Object Relations}\label{apx:object_relations}

Once the objects are selected, we compute all unary and binary relations between them using their continuous representations. We use a single feed-forward layer for unary and another for binary relations:

\begin{align}
  \operatorname{unary}(X, i)     & = \operatorname{tanh}(\mW_iX + b_i)         \\
  \operatorname{binary}(X, Y, j) & = \operatorname{tanh}(\mW_j[X,Y,X-Y] + b_j)
\end{align}

where $[]$ is the concatenation operator and $\mW, b$ are different weights for each equation. The result can be considered a fully connected graph with the selected objects as nodes, and the computed unary and binary relations as edges. We opt for this single-layer formulation to keep it computationally less intensive and focus on rule learning. We leave more complex and expressive relational layers that compute interactions between dense entity vectors such as Neural Tensor Networks~\cite{neuraltensornetworks} as future work.

\subsection{Image Reconstruction}\label{apx:image_reconstruction}
\begin{figure}
  \centering
  \includegraphics[width=0.9\textwidth]{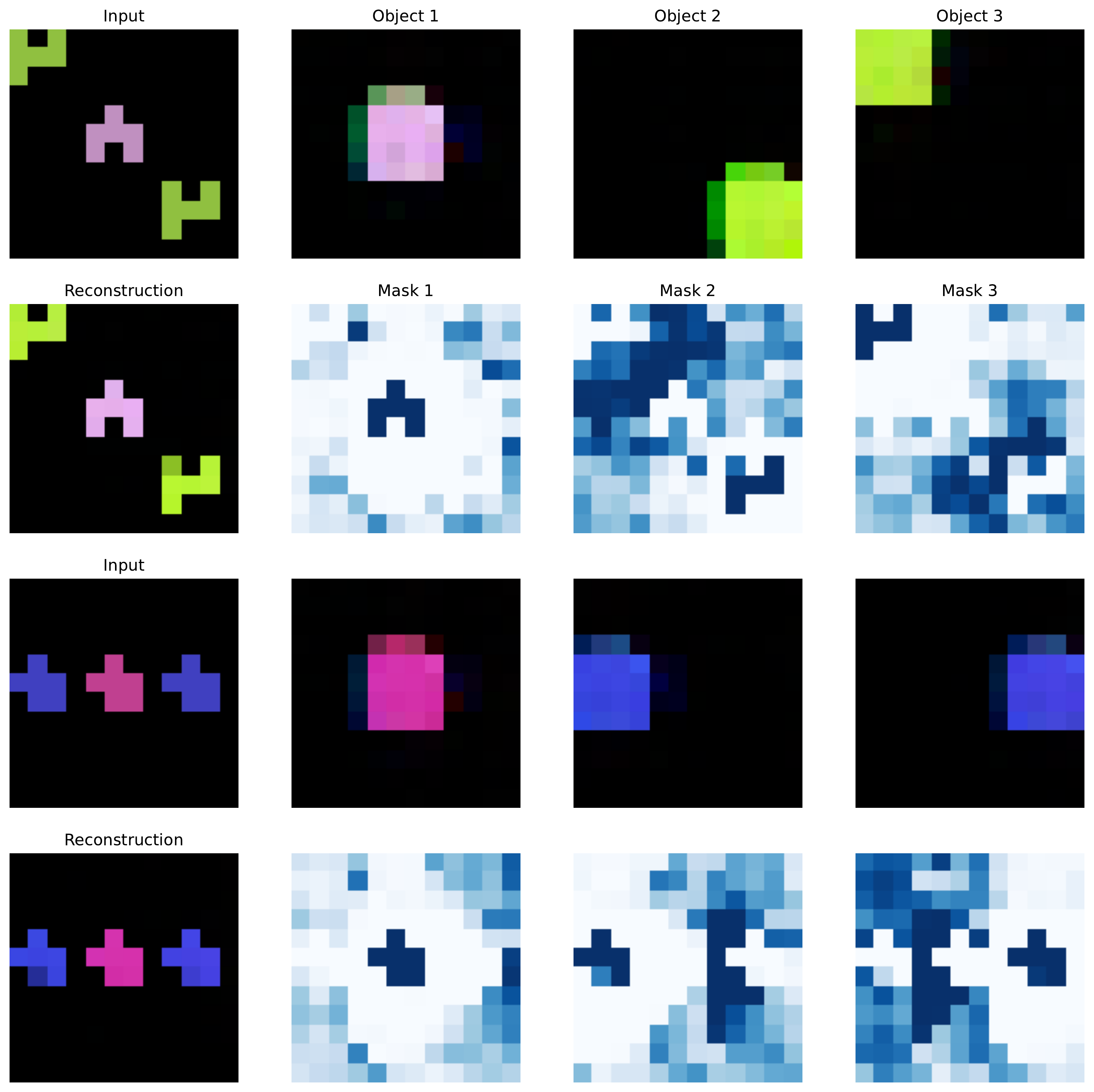}
  \caption{Example image reconstructions of the analysed DNF model with image reconstruction. The model seems to paint a colour and then use the mask like a cookie-cutter to extract out the shape.}
  \label{fig:relsgame_DNF-i_image_reconstructions}
\end{figure}

\begin{figure}
  \centering
  \includegraphics[width=0.9\textwidth]{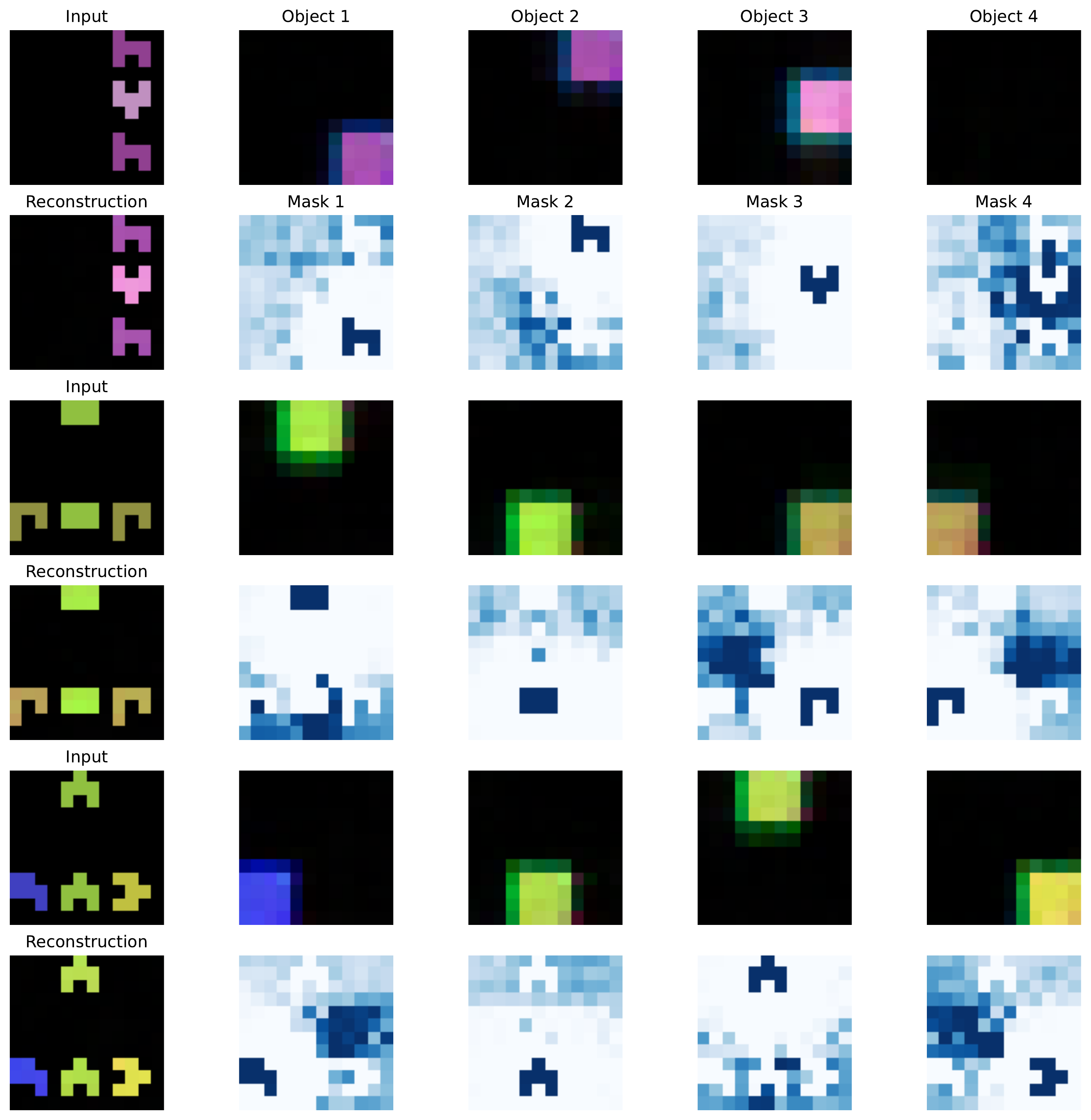}
  \caption{Further image reconstruction examples obtained from DNF-hi trained on all tasks with 1k examples per task.}
  \label{fig:relsgame_DNF-hi_image_reconstructions}
\end{figure}

To test the hypothesis whether image reconstruction can help with differentiable rule learning, we provide an auxiliary loss by reconstructing the full input image from the selected objects. Starting from a matrix of selected objects $\mO \in \R^{m \times d}$, we spatially broadcast them into a 3x3 grid. In the specific case of relations game dataset, the selected objects are broadcast into a tensor of shape mx3x3x36. We then use 2 deconvolution layers, also known as transposed convolution, with 32 filters, kernel size 5, relu activation and stride 2 to expand the tensor to shape mx12x12x32. For the final layer, we apply another deconvolution with 4 filters, kernel size 5 and a stride of 1 yielding mx12x12x4. The first three colour channels are combined with the last masking channel to obtain the final output.
The reconstructed image is trained using mean squared error and example reconstructions can be seen in \cref{fig:relsgame_DNF-i_image_reconstructions} and in \cref{fig:relsgame_DNF-hi_image_reconstructions}.

\subsection{Hyper-parameters}\label{apx:hyperparameters}
There are three main categories of hyperparameters used in this work: model, dataset and training. We cover all of them in \cref{tab:hyperparams_training} and \cref{tab:hyperparams_models}. The magnitude of the semantic gate selector $|\delta|$ presented in \cref{sec:sl_layer} is gradually adjusted during training according to a fixed exponential schedule. For the image classifier model, we start with 0.01 and increase to 1.0 with an exponential rate of 1.1 while for the subgraph set isomorphism dataset we start higher 0.1 and use the same rate of 1.1. We start the gate at a higher value for the graph isomorphism dataset since the input is already symbolic, i.e. $\forall_i x_i \in \{-1, 1\}$. The sign of $\delta$ is pre-determined by the layer type, conjunctive or disjunctive within a DNF layer, see \cref{sec:sl_layer,sec:experiments}.

\begin{table}
  \footnotesize
  \caption{All hyperparameters used for training the models.}
  \label{tab:hyperparams_training}
  \begin{tabular}{rcp{0.5\linewidth}}
    \toprule
    Key                           & Value & Comment                                                                                                                   \\
    \midrule
    \# batch updates Rels. Game   & 300k  & Number of training batch updates used.                                                                                    \\
    \# batch updates Graph        & 100k  &                                                                                                                           \\
    \# batch updates per eval     & 200   & How often to evaluate model on test dataset, number of batch updates divided by this quantity gives the number of epochs. \\
    learning rate                 & 0.001 & Optimiser learning rate, fixed throughout training.                                                                            \\
    batch size Rels. Game         & 64    & Batch size used for training.                                                                                             \\
    batch size Graph              & 128   &                                                                                                                           \\
    \# of repeated runs           & 5     & Every deep learning model on both datasets are trained 5 times with the same configuration.                               \\
    Input noise stddev Rels. Game & 0.01  & Noise added to input image in the relations game dataset.                                                                 \\
    rng seed Rels. Game           & 42    & Random number generator seed used in relations game data augmentation.                                                    \\
    rng seeds Graph               & .     & We use the 7 winning numbers of EuroMillions 25 December 2020.                                                            \\
    \bottomrule
  \end{tabular}
\end{table}

\begin{table}
  \scriptsize
  \caption{All hyperparameters used for constructing the models.}
  \label{tab:hyperparams_models}
  \begin{tabular}{rcp{0.5\linewidth}}
    \toprule
    Key                                  & Value & Comment                                                                                                                                                                                                                                                                                                                                                                  \\
    \midrule
    \# selected objects Same             & 2     & Number of objects selected in the Same task in the relations game dataset.                                                                                                                                                                                                                                                                                               \\
    \# variables Same                    & 2     & Number of variables used in the final DNF layer used to learn rules in the relations game dataset.                                                                                                                                                                                                                                                                       \\
    \# selected objects Between          & 3     &                                                                                                                                                                                                                                                                                                                                                                          \\
    \# variables Between                 & 3     &                                                                                                                                                                                                                                                                                                                                                                          \\
    \# selected objects Occurs           & 4     &                                                                                                                                                                                                                                                                                                                                                                          \\
    \# variables Occurs                  & 2     &                                                                                                                                                                                                                                                                                                                                                                          \\
    \# selected objects XOccurs          & 4     &                                                                                                                                                                                                                                                                                                                                                                          \\
    \# variables XOccurs                 & 4     &                                                                                                                                                                                                                                                                                                                                                                          \\
    \# selected objects All              & 4     &                                                                                                                                                                                                                                                                                                                                                                          \\
    \# variables All                     & 4     &                                                                                                                                                                                                                                                                                                                                                                          \\
    \# unary relations                   & 8     & Number of unary relations computed for the selected objects prior to DNF layer, the relations box in \cref{fig:relsgame_model}.                                                                                                                                                                                                                                              \\
    \# binary relations                  & 16    &                                                                                                                                                                                                                                                                                                                                                                          \\
    \# relations PrediNet                & 16    & Number of relations computed by the shared weights of each head.                                                                                                                                                                                                                                                                                                         \\
    \# heads PrediNet                    & .     & The number of heads is adjusted to match the output size of the relations computed prior to the DNF layer for fair comparison. Let $s,u,b$ be the number of selected objects, unary and binary relations used in the DNF models, then the number of heads is $h = su + s(s-1)b$. We adjust this so that the relation representation sizes are equal for a fairer comparison. \\
    PrediNet key size                    & 32    & The key size used for computing dot-product attention maps.                                                                                                                                                                                                                                                                                                              \\
    PrediNet output hidden size          & 64    & Size of the hidden layer of the output MLP used in PrediNet model.                                                                                                                                                                                                                                                                                                       \\
    input CNN hidden size                & 32    & The number of filters used by the input CNN for both DNF and PrediNet models.                                                                                                                                                                                                                                                                                            \\
    input CNN activation                 & relu  &                                                                                                                                                                                                                                                                                                                                                                          \\
    \# variables hidden DNF              & 2     & Number of variables that can appear in the rules learnt by the hidden DNF layer of the DNF-h models.                                                                                                                                                                                                                                                                     \\
    \# rule definitions hidden DNF       & 4     & Maximum number of rule definitions per predicate in the hidden layer.                                                                                                                                                                                                                                                                                                    \\
    \# invented predicates hidden DNF    & 14    & The number of predicates learnt by the hidden DNF layer. Specifically, 2 nullary, 4 unary and 8 binary predicates.                                                                                                                                                                                                                                                       \\
    \# rule definitions DNF Rels. Game   & 8     & The target label can be defined at most by 8 different rules.                                                                                                                                                                                                                                                                                                            \\
    \# invented predicates recursive DNF & 7     & The number of extra predicates learnt by the DNF layer in the recursive configuration DNF-r. Specifically, 1 nullary, 2 unary and 4 binary predicates.                                                                                                                                                                                                                   \\
    \# iterations recursive DNF          & 2     &                                                                                                                                                                                                                                                                                                                                                                          \\
    \# rule definitions recursive DNF    & 2     & Maximum number of rule definitions per predicate including the label in the recursive DNF layer, DNF-r.                                                                                                                                                                                                                                                                  \\
    \bottomrule
  \end{tabular}
\end{table}

\section{Dataset Details}\label{apx:dataset_details}
We use two synthetic datasets to create a controlled environment in which we can evaluate our approach. The first dataset is based on the subgraph isomorphism problem and is used as the basis for an unbiased Inductive Logic Programming (ILP)~\cite{ilp} task. The second dataset is an image classification task which involves objects of different shapes and colour in some compound relation such as between two identical objects.

\subsection{Subgraph Set Isomorphism}\label{apx:gendnf_details}

The subgraph set isomorphism task is designed to create an unbiased combinatorial search space for rule learning. To generate data points, we uniformly sample unique graphs for $\sH$ and then sample $\gG$ of which subgraphs are isomorphic. The resulting examples are checked using the answer set solver clingo~\cite{clingo5}. In total we sample 10k graphs and take only the unique ones to ensure any partitions of the data are disjoint. We then split into training, validation and test with sizes 2k, 1k and 1k respectively.
Since each relation could be positively, negatively or be absent in the rule, the upper bound on the search space for possible rules is $3^{|E(\gH_i)|}$. This ensures there is no bias in the rules or heuristics that can be used to prune the search space. As a result, any atom can appear in the rule equally likely.

\begin{table}
  \centering
  \caption{Example target rule generated for the medium dataset size. This is the task that FastLAS does not terminate after 16 hours. Note that obj() predicate as well as the uniqueness of variables are added for ASP safe representation.}
  \label{tab:gendnf_medium}
  \footnotesize
  \begin{tabular}{rl}
    \toprule
    t :- & nullary(0), nullary(2), not nullary(3), unary(V0,0), unary(V0,1), not unary(V0,2),       \\
         & not unary(V0,3), not unary(V1,0), not unary(V1,1), not unary(V1,2), not unary(V1,3),     \\
         & unary(V2,0), unary(V2,2), binary(V0,V1,0), binary(V0,V1,1), binary(V0,V1,2),             \\
         & binary(V0,V1,3), not binary(V0,V1,4), not binary(V0,V1,5), not binary(V0,V2,0),          \\
         & binary(V0,V2,2), not binary(V0,V2,3), not binary(V0,V2,5), binary(V1,V0,1),              \\
         & not binary(V1,V0,2), not binary(V1,V0,3), not binary(V1,V0,4), not binary(V1,V0,5),      \\
         & binary(V1,V2,0), binary(V1,V2,3), not binary(V1,V2,4), binary(V2,V0,0),                  \\
         & not binary(V2,V0,2), binary(V2,V0,3), not binary(V2,V0,4), not binary(V2,V1,1),          \\
         & not binary(V2,V1,3), not binary(V2,V1,4), not binary(V2,V1,5), obj(V2), V2 != V0,        \\
         & V2 != V1, obj(V0), V0 != V1, obj(V1).                                                    \\
    t :- & nullary(0), not nullary(1), not nullary(2), unary(V0,0), unary(V0,1), unary(V0,3),       \\
         & unary(V0,4), not unary(V1,0), not unary(V1,2), not unary(V1,4), not unary(V2,0),         \\
         & not unary(V2,1), unary(V2,2), not unary(V2,3), not unary(V2,4), not binary(V0,V1,2),     \\
         & binary(V0,V1,3), binary(V0,V1,4), binary(V0,V1,5), not binary(V0,V2,1), binary(V0,V2,2), \\
         & binary(V0,V2,3), not binary(V0,V2,4), binary(V0,V2,5), not binary(V1,V0,2),              \\
         & binary(V1,V0,3), not binary(V1,V0,4), not binary(V1,V0,5), not binary(V1,V2,4),          \\
         & not binary(V1,V2,5), not binary(V2,V0,0), not binary(V2,V0,1), not binary(V2,V0,2),      \\
         & binary(V2,V0,4), not binary(V2,V1,0), not binary(V2,V1,1), not binary(V2,V1,4),          \\
         & obj(V2), V2 != V0, V2 != V1, obj(V0), V0 != V1, obj(V1).                                 \\
    \bottomrule
  \end{tabular}
\end{table}

An example set of rules $\sH$ from the medium dataset size is shown in \cref{tab:gendnf_medium}. The head of the rules $t$ corresponds to the desired target label for a given set of context facts, i.e. whether the given graph is subgraph isomorphic to the set of graphs represented by the rules. The symbolic learners timeout on this size due to the long rules. Rules of this length are uncommon in many standard ILP datasets~\cite{deltailp} used to evaluate a learner. In order for the rules to be safe in ASP, that is every variable appears in at least one positive atom, we add obj(...) predicate for every variable that is true for every grounded object. We also add the uniqueness of variables constraint by encoding $V_i \neq V_j$ if $i \neq j$ since in graph isomorphism every node can at most be mapped to one other node, i.e. a one-to-one mapping is required. This is only done when evaluating with clingo~\cite{clingo5} or using symbolic learners.

\subsection{Relations Game}\label{apx:relsgame_details}
\begin{figure}
  \centering
  \begin{subfigure}[b]{1.0\textwidth}
    \centering
    \includegraphics[width=0.7\textwidth]{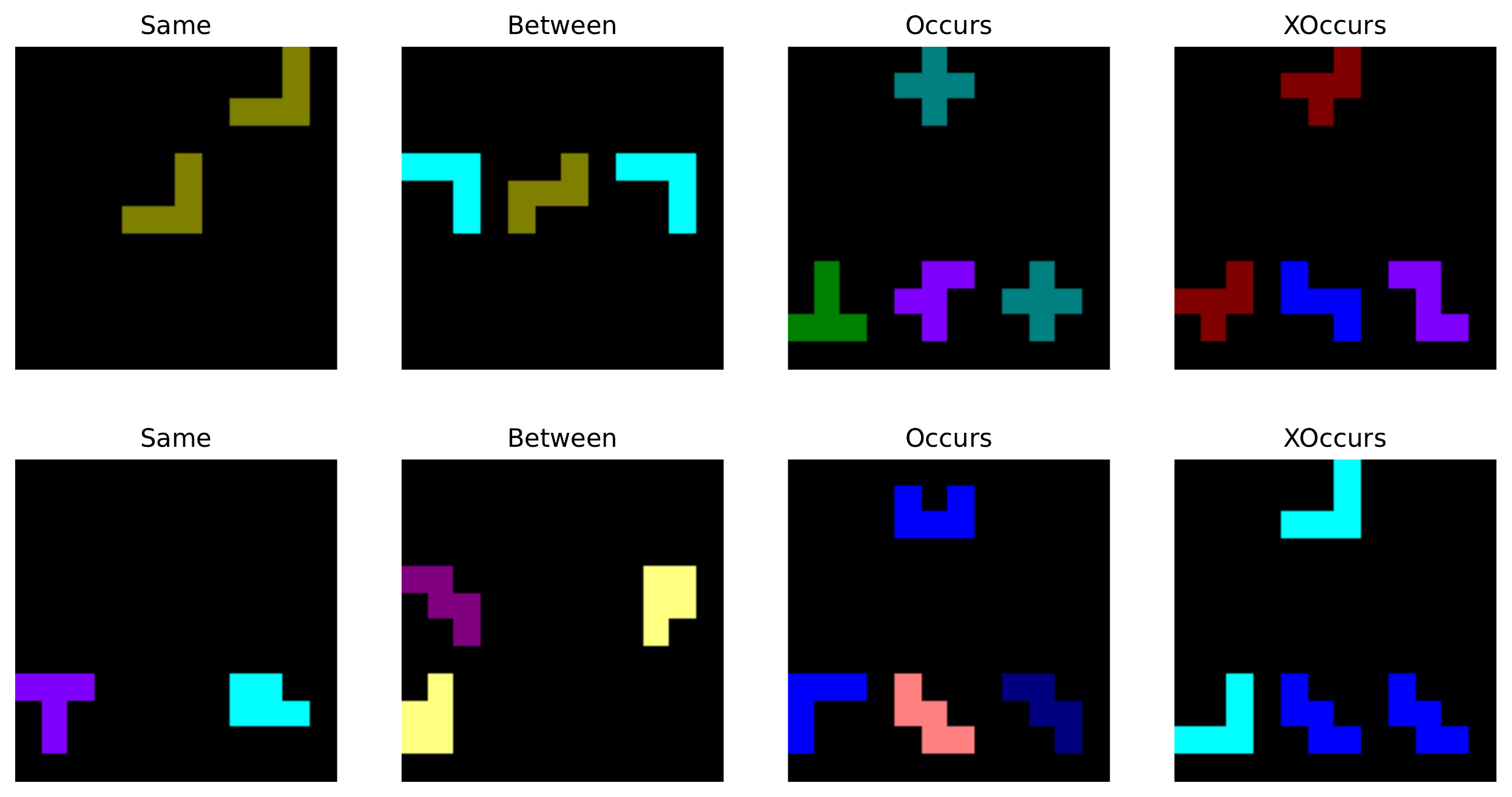}
    \caption{Example pentominoes where each shape consists of 5 pixels organised in a 3x3 grid.}
    \label{fig:relsgame_tasks_pentos}
  \end{subfigure}
  \begin{subfigure}[b]{1.0\textwidth}
    \centering
    \includegraphics[width=0.7\textwidth]{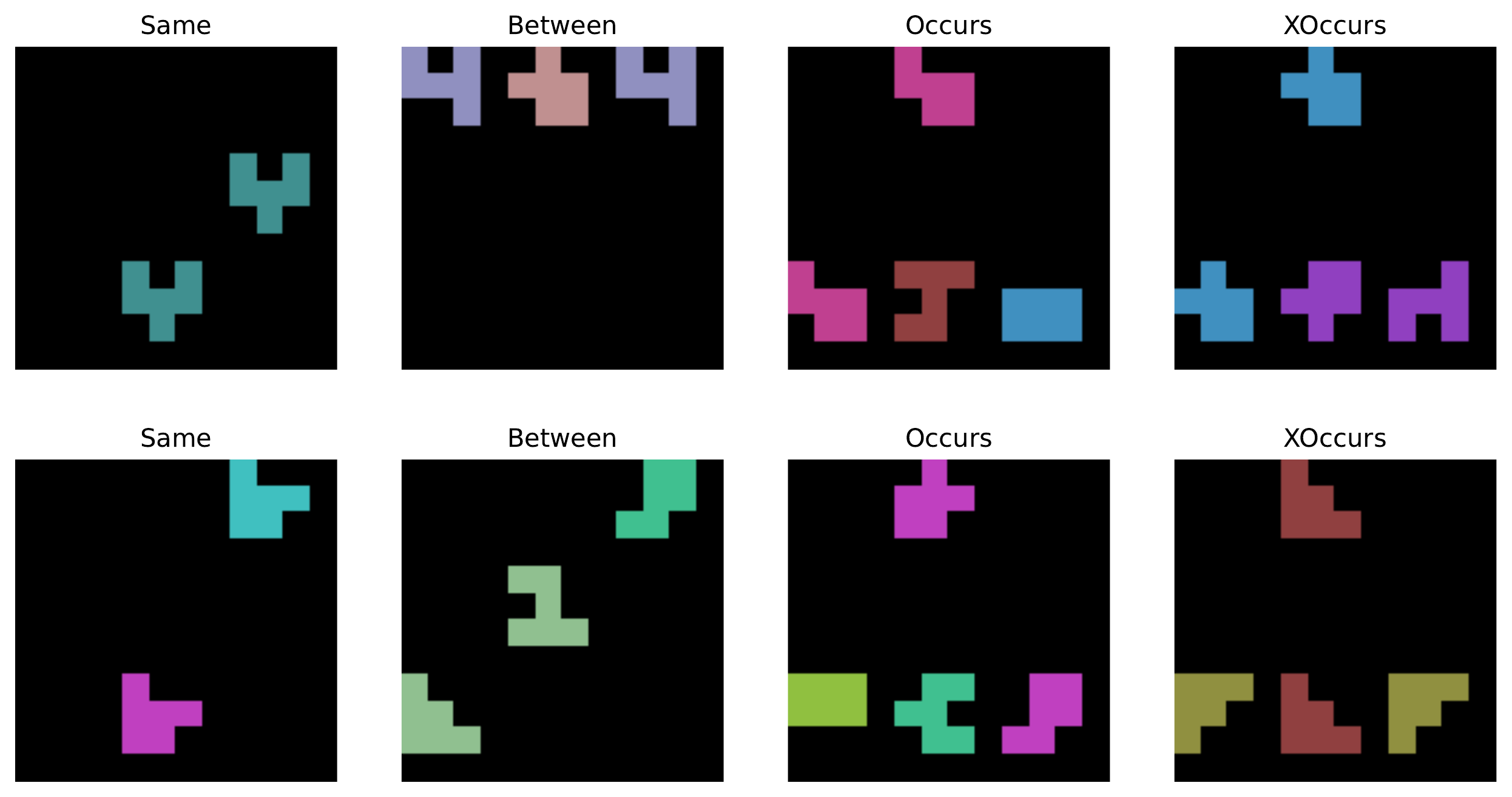}
    \caption{Example hexominoes where each shape consists of 6 pixels with unseen colours to pentominoes.}
    \label{fig:relsgame_tasks_hexos}
  \end{subfigure}
  \begin{subfigure}[b]{1.0\textwidth}
    \centering
    \includegraphics[width=0.7\textwidth]{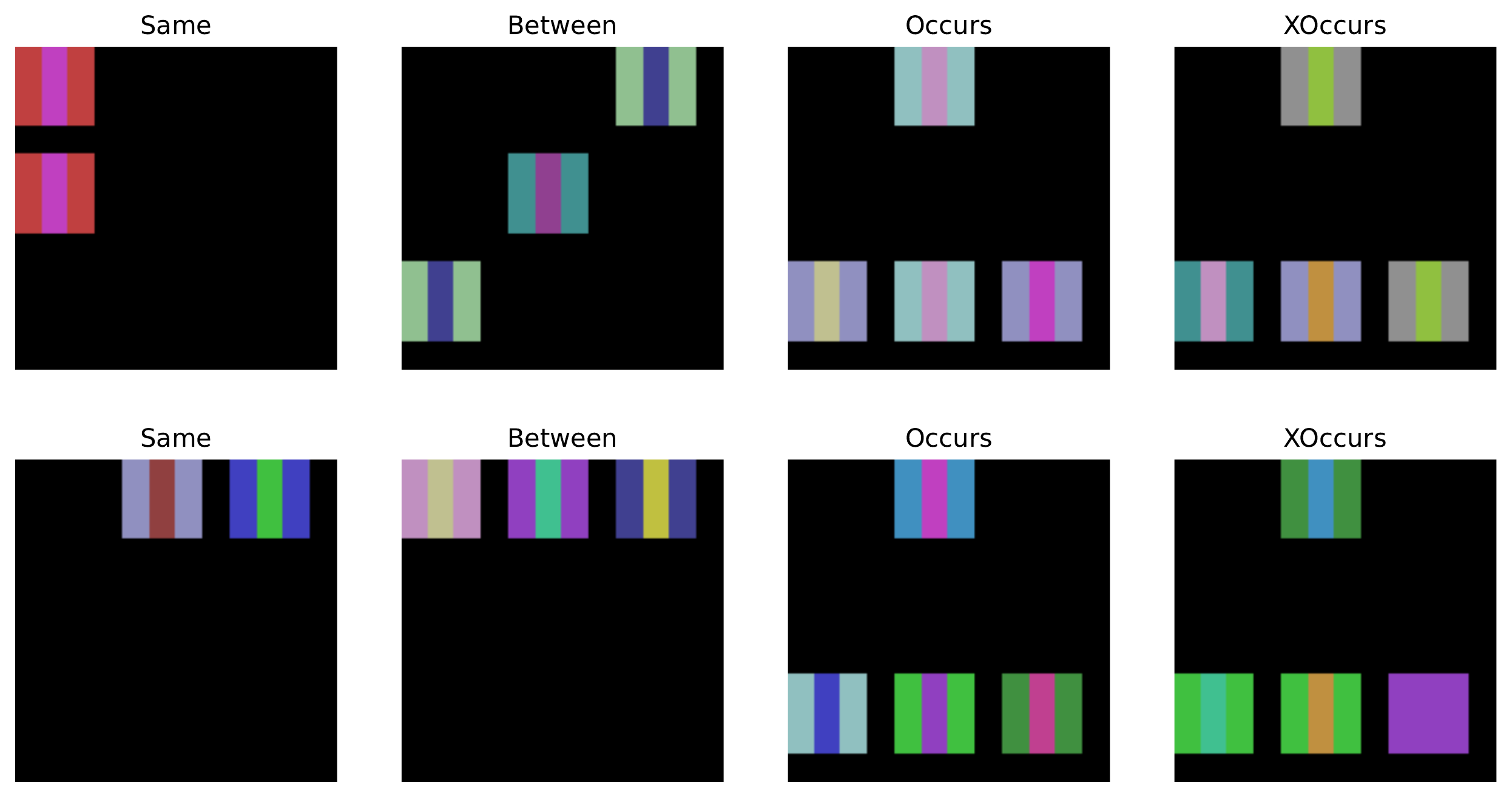}
    \caption{Example striped shapes where each object has 9 pixels with a striped colour pattern.}
    \label{fig:relsgame_tasks_stripes}
  \end{subfigure}
  \caption{Further samples from the relations game dataset. The top and bottom rows are for true and false cases respectively.}
  \label{fig:relsgame_set_samples}
\end{figure}

The relations game dataset consists of an input image with a desired binary label. There are 3 sets: pentominoes, hexominoes and stripes, examples of which are shown in \cref{fig:relsgame_set_samples}. The original dataset presented in PrediNet~\cite{predinet} comes with 250k examples per set and with a higher resolution of 36x36x3. At the high resolution, each \emph{block} of a shape consists of 3x3 pixels and since this is redundant, we convert each block to a single pixel reducing the image without any loss of information to 12x12x3. This conversion reduces the computational power required to run the experiments without changing the tasks.

During training we apply standard data augmentation: (i) random horizontal or vertical flips, (ii) random 90 degrees counter-clockwise rotation and (iii) added input noise drawn from a normal distribution with 0.01 as the standard deviation. This augmentation is applied to every batch and the randomness is drawn from a random number generator with a fixed seed of 42.

\section{Training Details}\label{apx:training_details}
We train all deep models using the Adam~\cite{adam} optimiser with a learning rate of 0.001. The models are trained for a fixed number of batch updates and evaluated every 200 batch updates. We use the negative log-likelihood as the loss function for the binary predictions. For all the hyperparameters used in training, please refer to \cref{tab:hyperparams_training}. The models are trained on a shared pool of computers all having Intel Core i7 CPUs. Note that the run times of the deep models may vary depending on the shared workload on the worker machine. Once the training is complete for the DNF models, we prune and threshold the weights to obtain DNF+t variants, as described in \cref{sec:sl_layer}.

\begin{figure}
  \centering
  \includegraphics[width=1.0\textwidth]{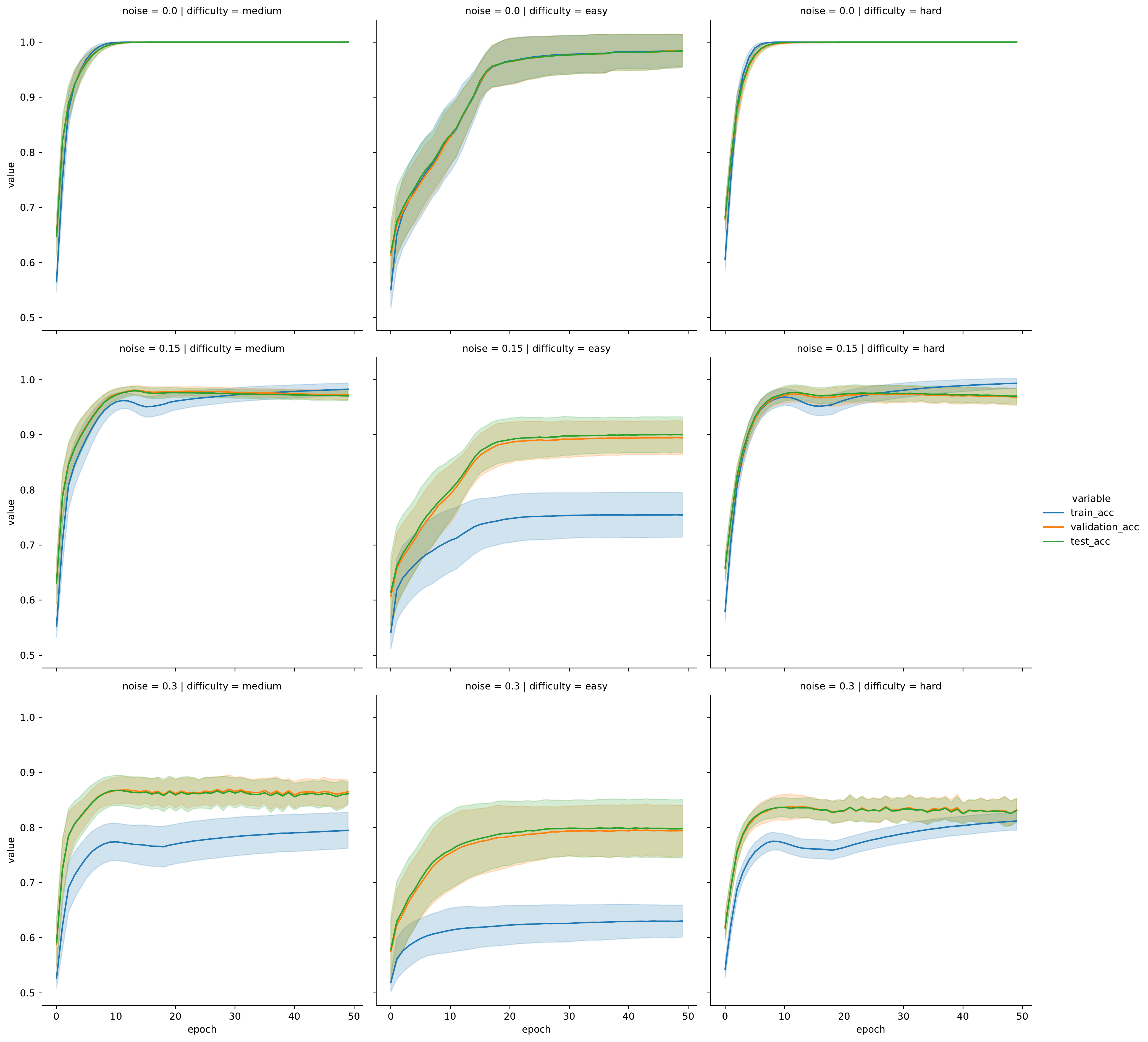}
  \caption{Training curves for DNF layer on the subgraph set isomorphism task. The model is training for 100k batch updates logging every 200 steps. This gives a total of 50 epochs shown on the x axis.}
  \label{fig:gendnf_training_curves}
\end{figure}

\begin{figure}
  \centering
  \includegraphics[width=0.9\textwidth]{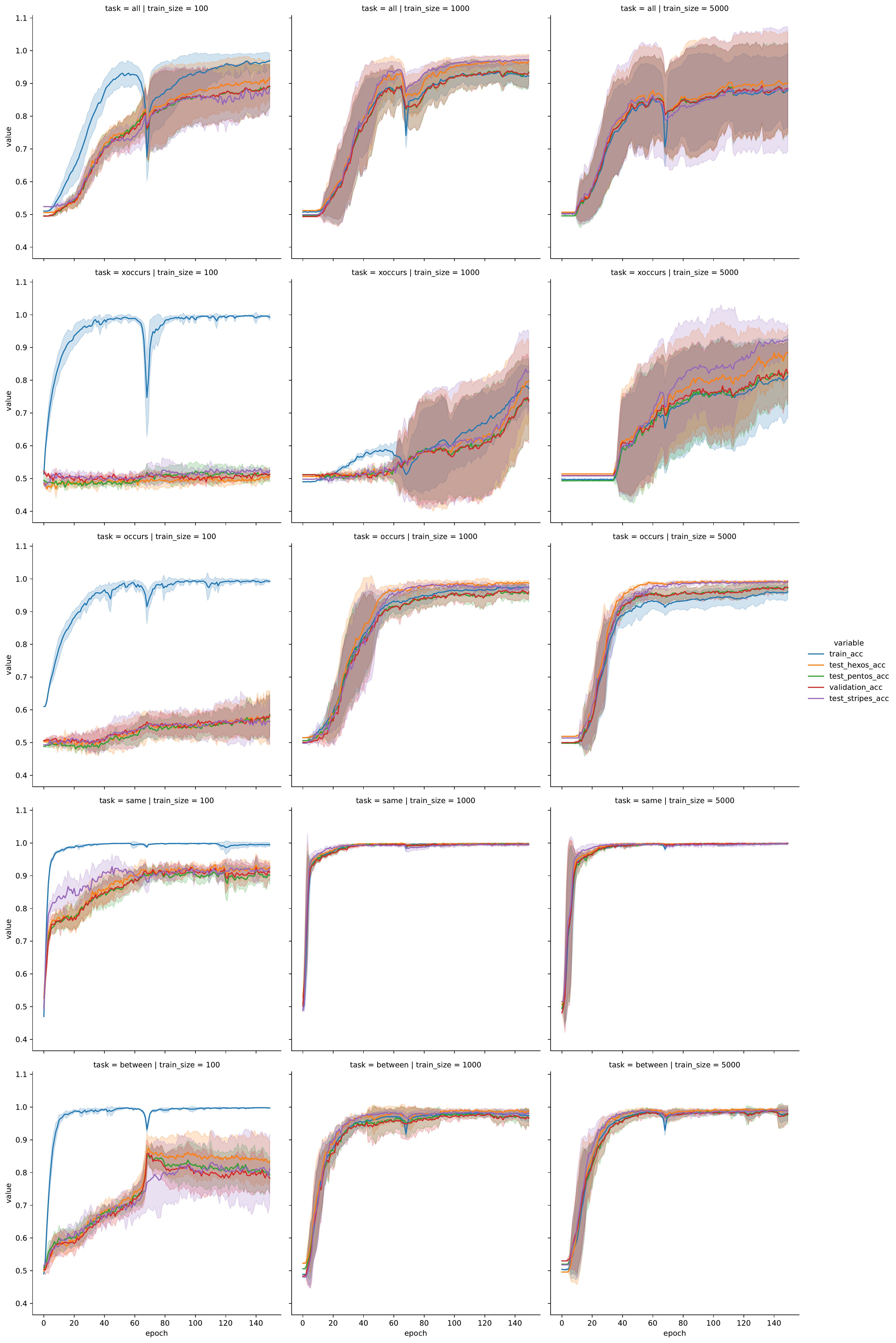}
  \caption{Training curves for the DNF model relations game task.}
  \label{fig:relsgame_dnf_training_curves}
\end{figure}

\begin{figure}
  \centering
  \includegraphics[width=0.9\textwidth]{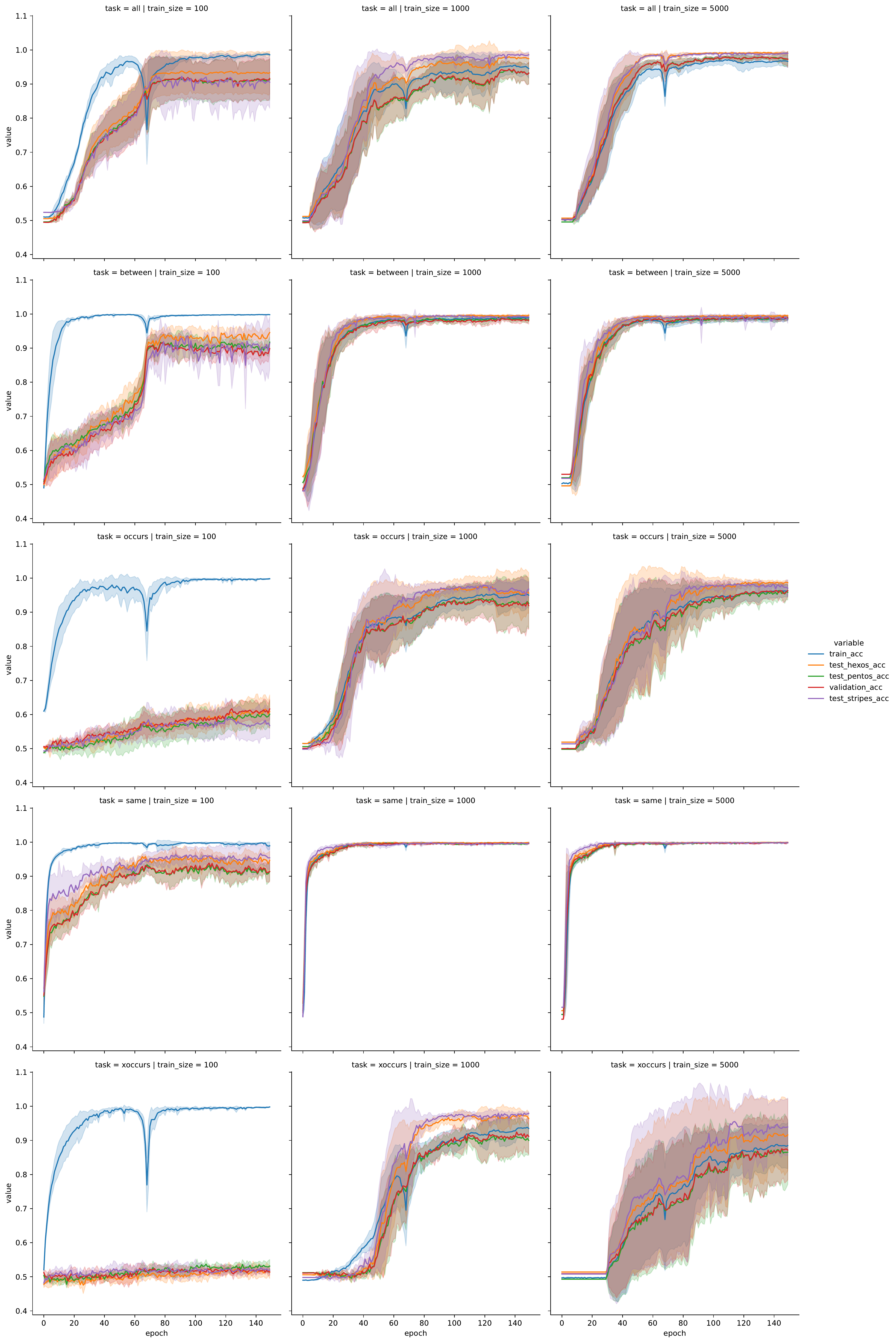}
  \caption{Training curves for the DNF model with a hidden layer (DNF-h) on the relations game tasks.}
  \label{fig:relsgame_dnf-h_training_curves}
\end{figure}

\begin{figure}
  \centering
  \includegraphics[width=0.9\textwidth]{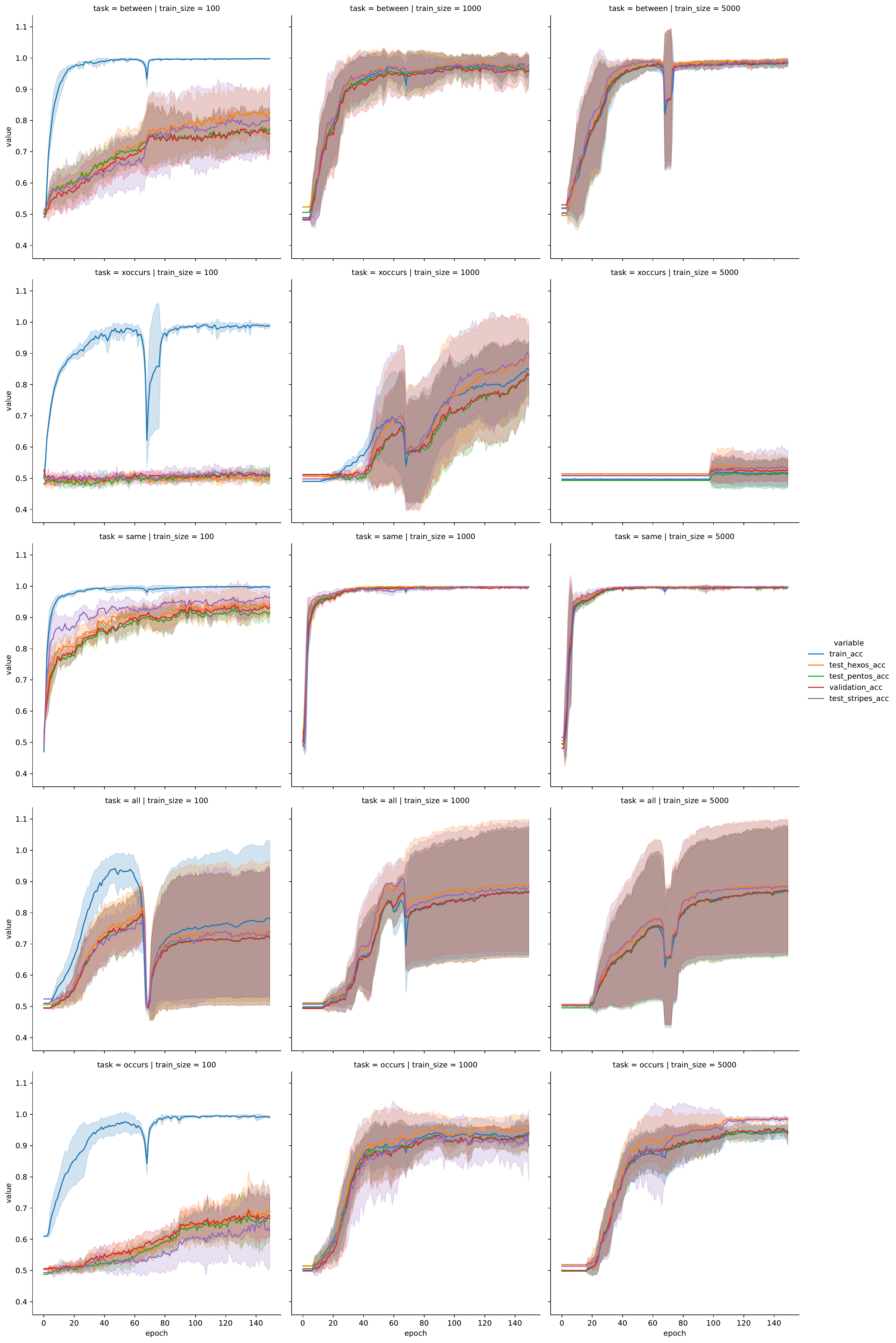}
  \caption{Training curves for the recursive DNF model (DNF-r) on the relations game tasks.}
  \label{fig:relsgame_dnf-r_training_curves}
\end{figure}

\begin{figure}
  \centering
  \includegraphics[width=0.9\textwidth]{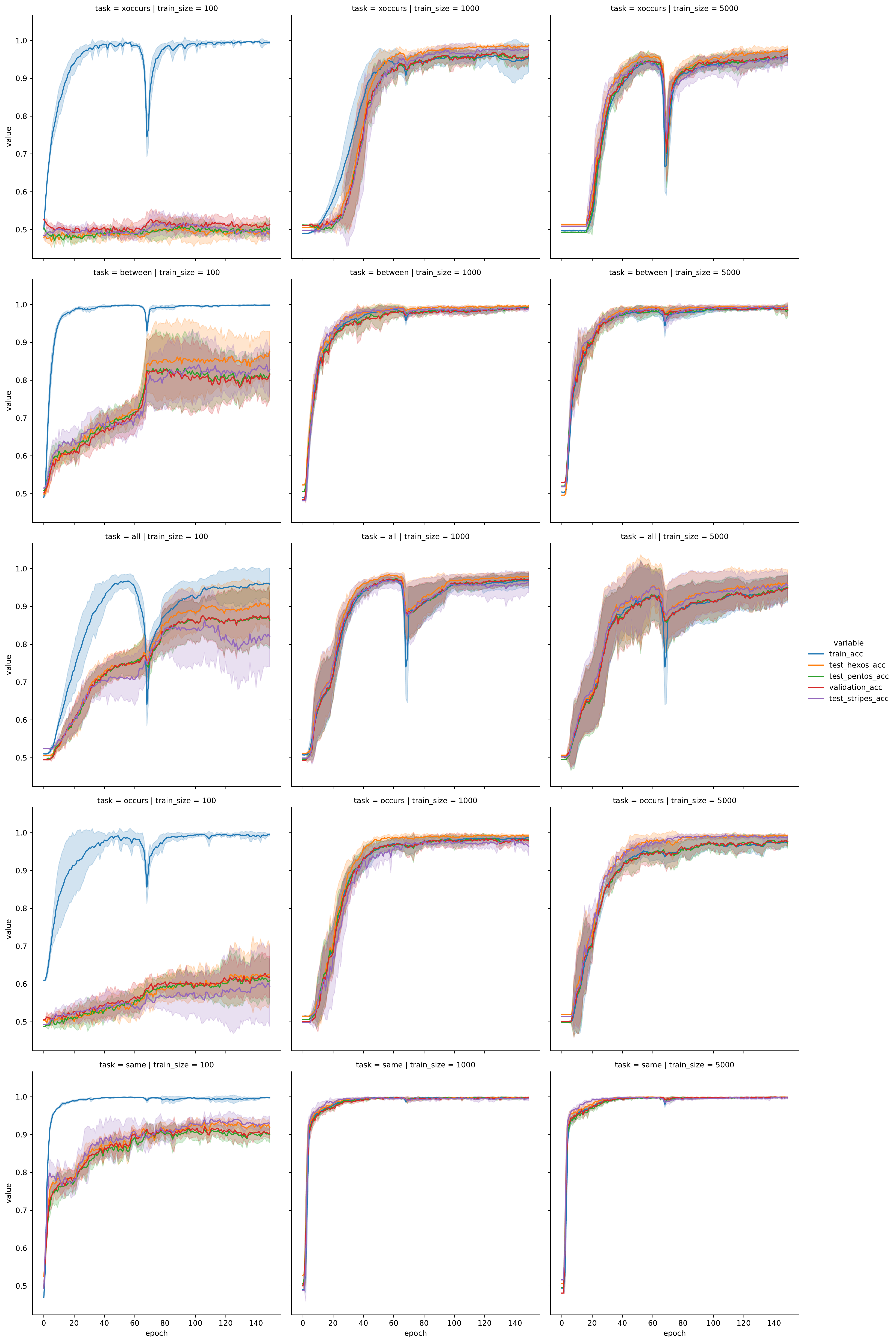}
  \caption{Training curves for the DNF model with image reconstruction loss (DNF-i).}
  \label{fig:relsgame_dnf-i_training_curves}
\end{figure}

\begin{figure}
  \centering
  \includegraphics[width=0.8\textwidth]{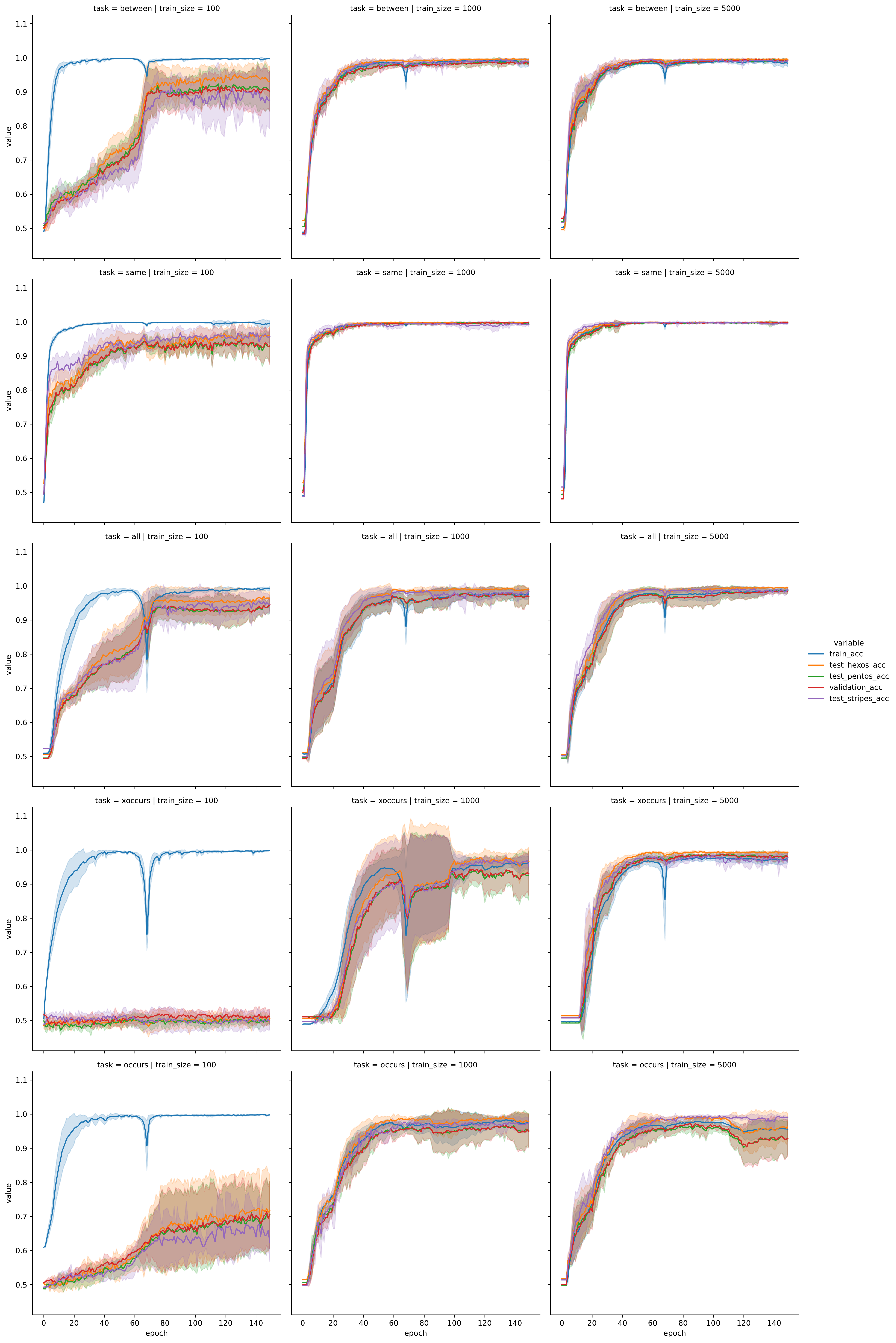}
  \caption{Training curves for the DNF model with a hidden layer (DNF-hi) with image reconstruction loss.}
  \label{fig:relsgame_dnf-hi_training_curves}
\end{figure}

\begin{figure}
  \centering
  \includegraphics[width=0.9\textwidth]{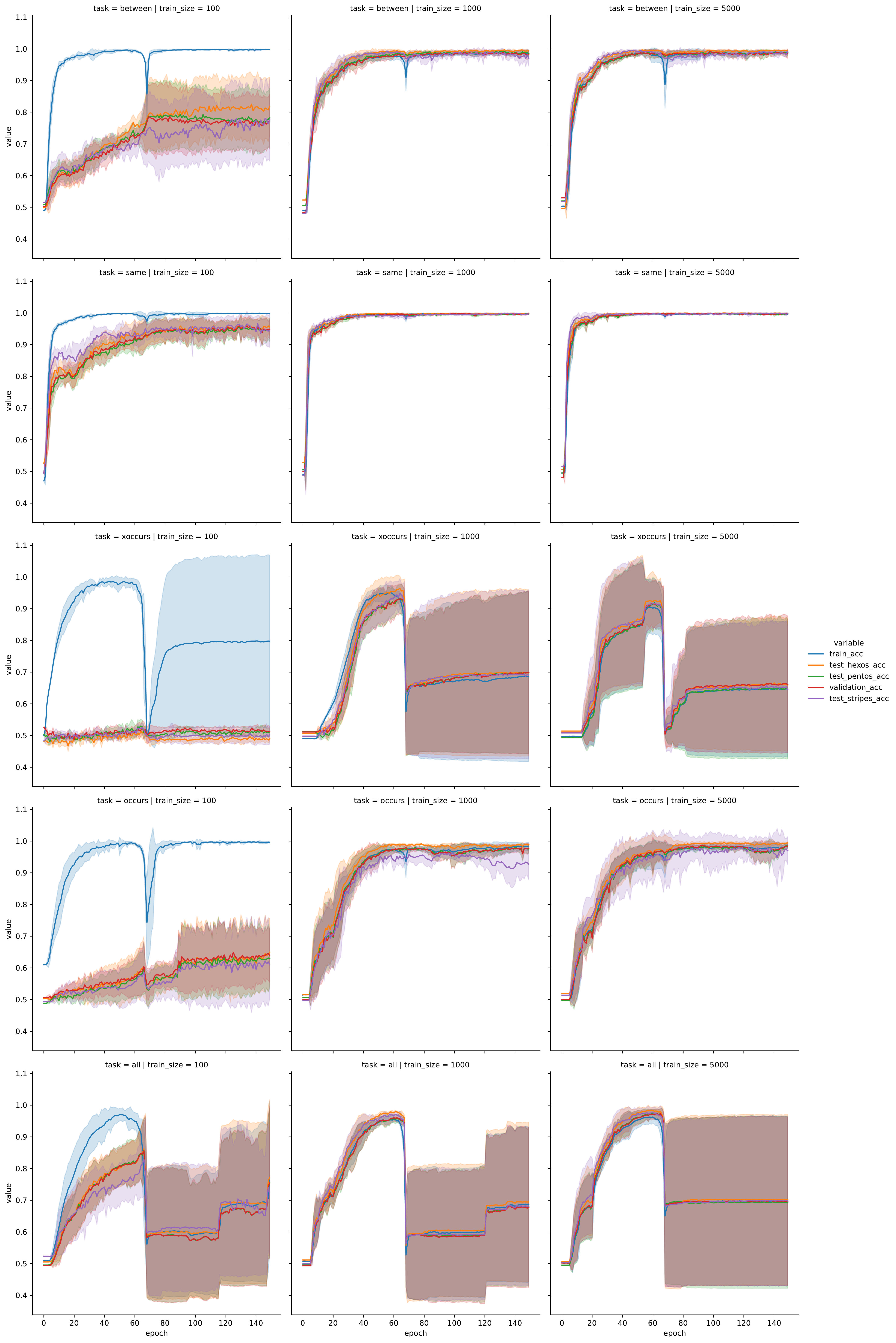}
  \caption{Training curves for the recursive DNF model with reconstruction loss (DNF-ri).}
  \label{fig:relsgame_dnf-ri_training_curves}
\end{figure}

\begin{figure}
  \centering
  \includegraphics[width=0.9\textwidth]{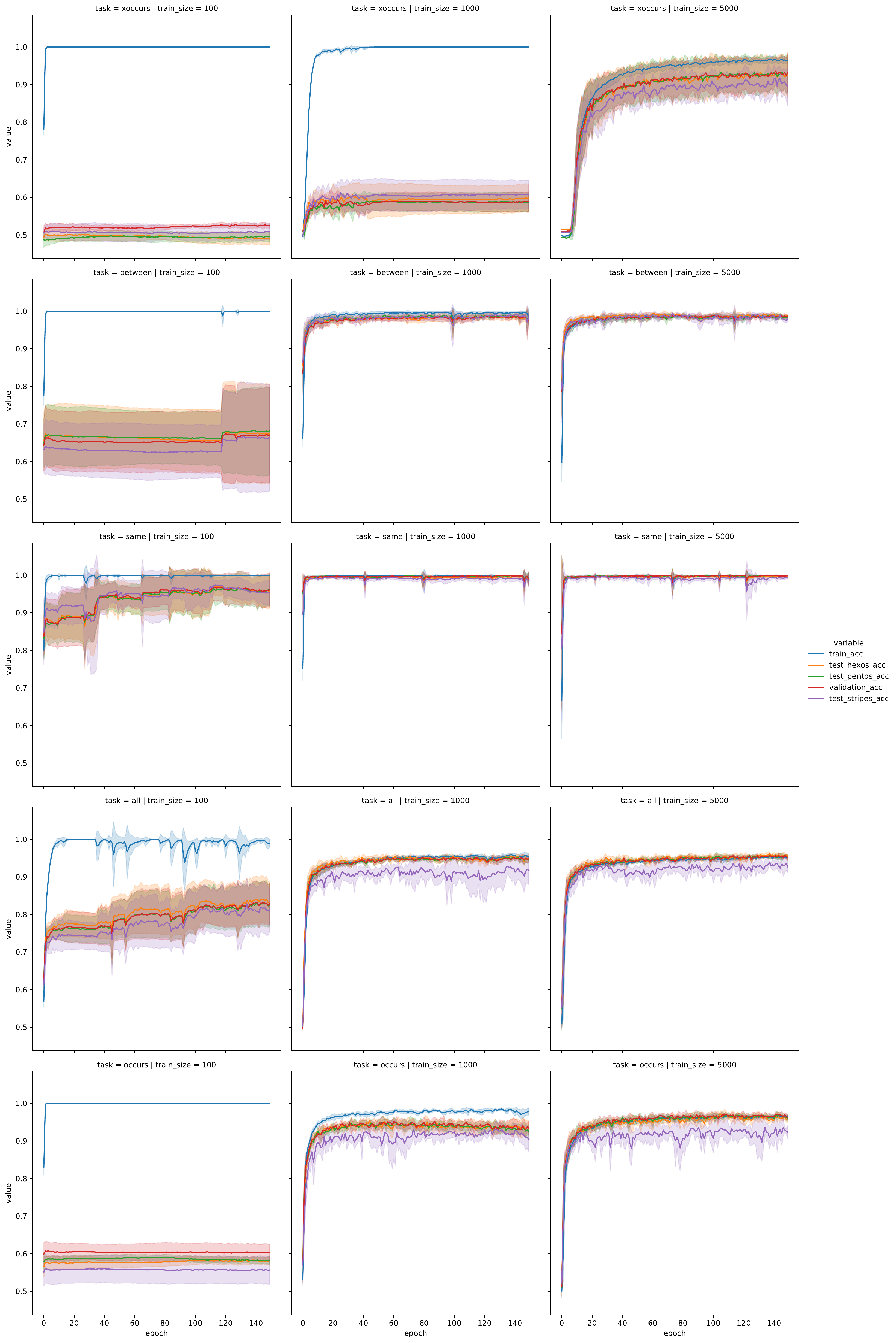}
  \caption{Training curves for PrediNet on relations game tasks.}
  \label{fig:relsgame_predinet_training_curves}
\end{figure}

The training curves for every deep model trained on every dataset and hyperparameter configuration, can be found in \cref{fig:gendnf_training_curves}, \cref{fig:relsgame_dnf_training_curves}, \cref{fig:relsgame_dnf-h_training_curves}, \cref{fig:relsgame_dnf-r_training_curves}, \cref{fig:relsgame_dnf-i_training_curves}, \cref{fig:relsgame_dnf-hi_training_curves}, \cref{fig:relsgame_dnf-ri_training_curves} and \cref{fig:relsgame_predinet_training_curves}. For relations game training curves, the slight dip in accuracy around epoch 70 when trained with only 100 examples corresponds to the point at which the semantic gate $|\delta|$ becomes close to 1. This indicates that if the model is overfitting, it almost has to relearn the task with the desired semantics since this phenomenon is less pronounced with more training examples and almost absent in the graph dataset. We also observe mode collapses in the recursive configuration of the DNF model (DNF-r) in which the accuracy suddenly drops when the semantic gate $|\delta|$ becomes closer to 1 around epoch 70. We believe this is due to the recursion in the model making a recovery to the desired disjunctive normal form semantics more difficult. The DNF-r either performs well with some runs surviving the saturation $|\delta|=1.0$ and some collapsing.

\section{Further Results}\label{apx:further_results}
This section includes further tables and figures to complement and extend the main content presented in the paper. The description for each figure is provided in the caption.

\begin{figure}
  \centering
  \includegraphics[width=1.0\textwidth]{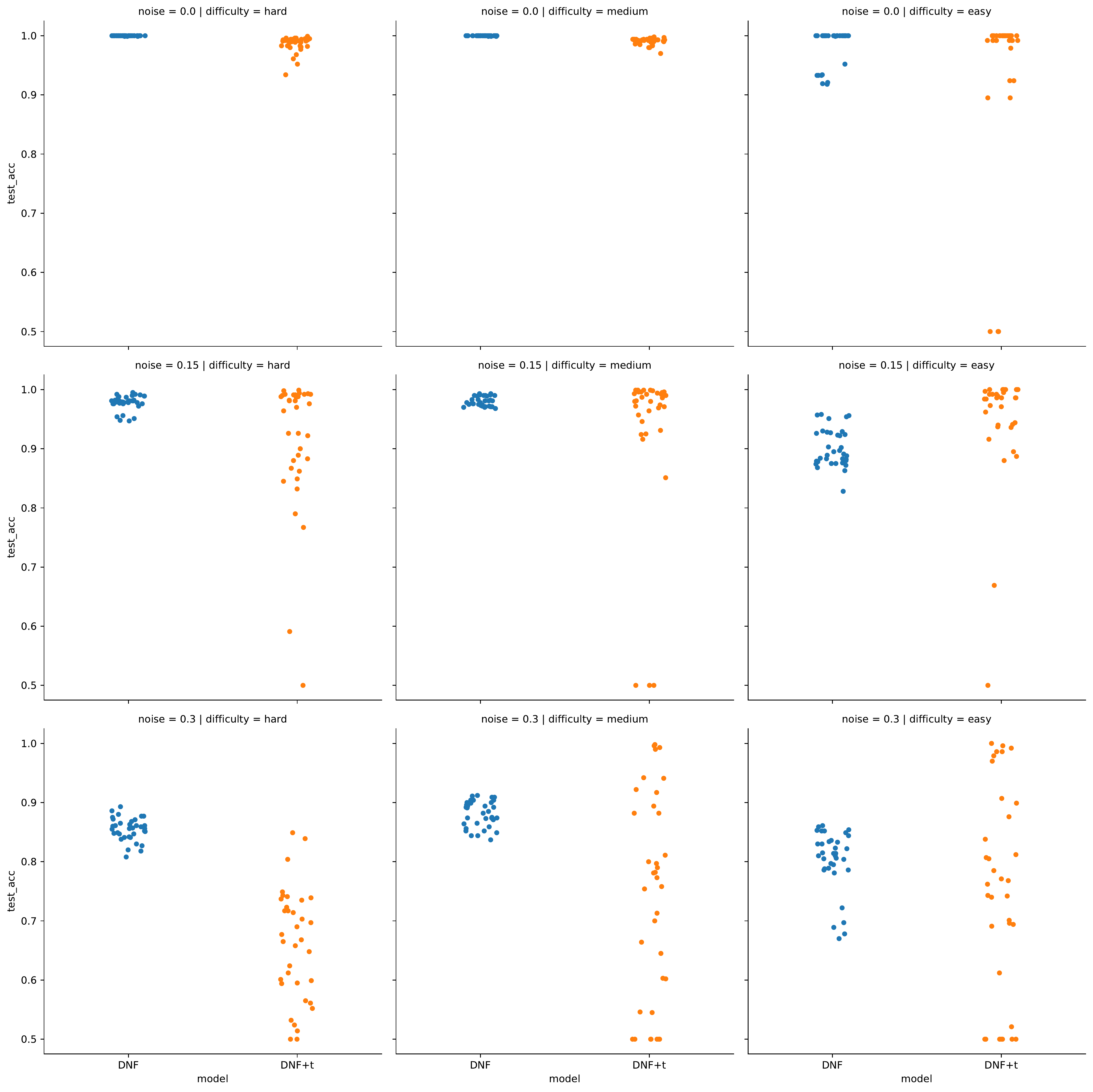}
  \caption{All the results for DNF layer on the subgraph set isomorphism task. There 315 unique runs: 5 runs for each 7 seeds for every difficulty (3) with 3 levels of noise. For each run, the thresholded results are shown as DNF+t.}
  \label{fig:all_dnf_data}
\end{figure}

\begin{figure}
  \centering
  \includegraphics[width=1.0\textwidth]{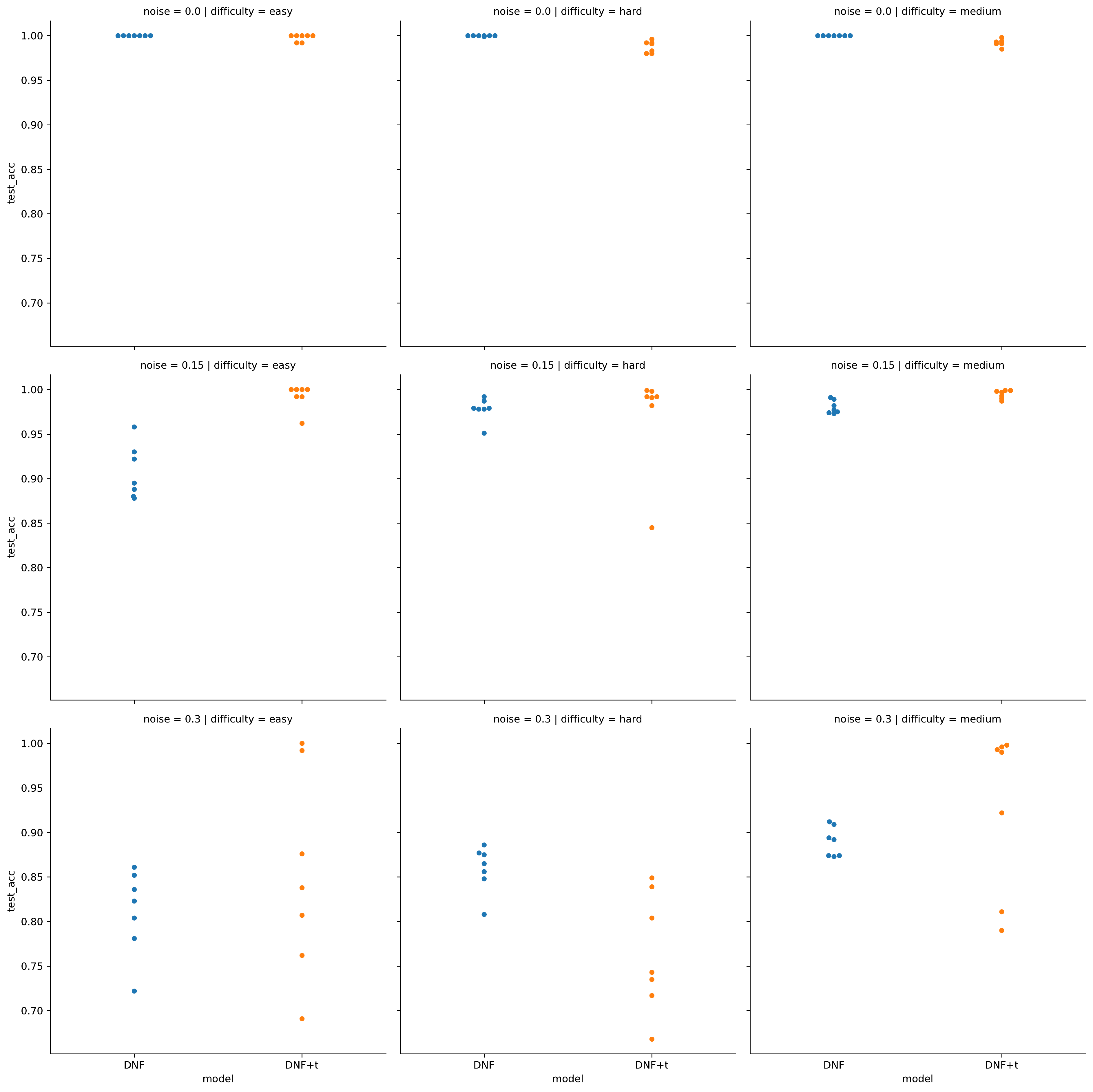}
  \caption{Best results filtered out of 5 runs from \cref{fig:all_dnf_data}. There is one point for every random seed used, 7 in total. These are all the data points used in \cref{tab:best_dnf_full}.}
  \label{fig:best_dnf_data}
\end{figure}

\begin{table}
  \centering
  \caption{Full results of the best runs of the DNF layer on the subgraph set isomorphism task.}
  \label{tab:best_dnf_full}
  \begin{tabular}{llrrrrrrrr}
    \toprule
               & {}    & \multicolumn{4}{l}{median}    & \multicolumn{4}{l}{mad}                                                                                                                   \\
               & {}    & \multicolumn{2}{l}{test acc} & \multicolumn{2}{l}{validation acc} & \multicolumn{2}{l}{test acc} & \multicolumn{2}{l}{validation acc}                                 \\
               & model & DNF                           & DNF+t                               & DNF                           & DNF+t                               & DNF   & DNF+t & DNF   & DNF+t \\
    difficulty & noise &                               &                                     &                               &                                     &       &       &       &       \\
    \midrule
    easy       & 0.00  & 1.000                         & 1.000                               & 1.000                         & 1.000                               & 0.000 & 0.003 & 0.000 & 0.002 \\
               & 0.15  & 0.895                         & 1.000                               & 0.909                         & 1.000                               & 0.025 & 0.009 & 0.022 & 0.010 \\
               & 0.30  & 0.823                         & 0.838                               & 0.809                         & 0.827                               & 0.036 & 0.089 & 0.030 & 0.085 \\
    hard       & 0.00  & 1.000                         & 0.991                               & 1.000                         & 1.000                               & 0.000 & 0.006 & 0.000 & 0.000 \\
               & 0.15  & 0.979                         & 0.992                               & 0.981                         & 0.995                               & 0.008 & 0.036 & 0.008 & 0.034 \\
               & 0.30  & 0.865                         & 0.743                               & 0.877                         & 0.769                               & 0.019 & 0.056 & 0.012 & 0.044 \\
    medium     & 0.00  & 1.000                         & 0.993                               & 1.000                         & 1.000                               & 0.000 & 0.003 & 0.000 & 0.000 \\
               & 0.15  & 0.977                         & 0.997                               & 0.983                         & 1.000                               & 0.006 & 0.004 & 0.005 & 0.002 \\
               & 0.30  & 0.892                         & 0.990                               & 0.902                         & 0.997                               & 0.014 & 0.075 & 0.017 & 0.074 \\
    \bottomrule
  \end{tabular}
\end{table}

\begin{figure}
  \centering
  \includegraphics[width=1.0\textwidth]{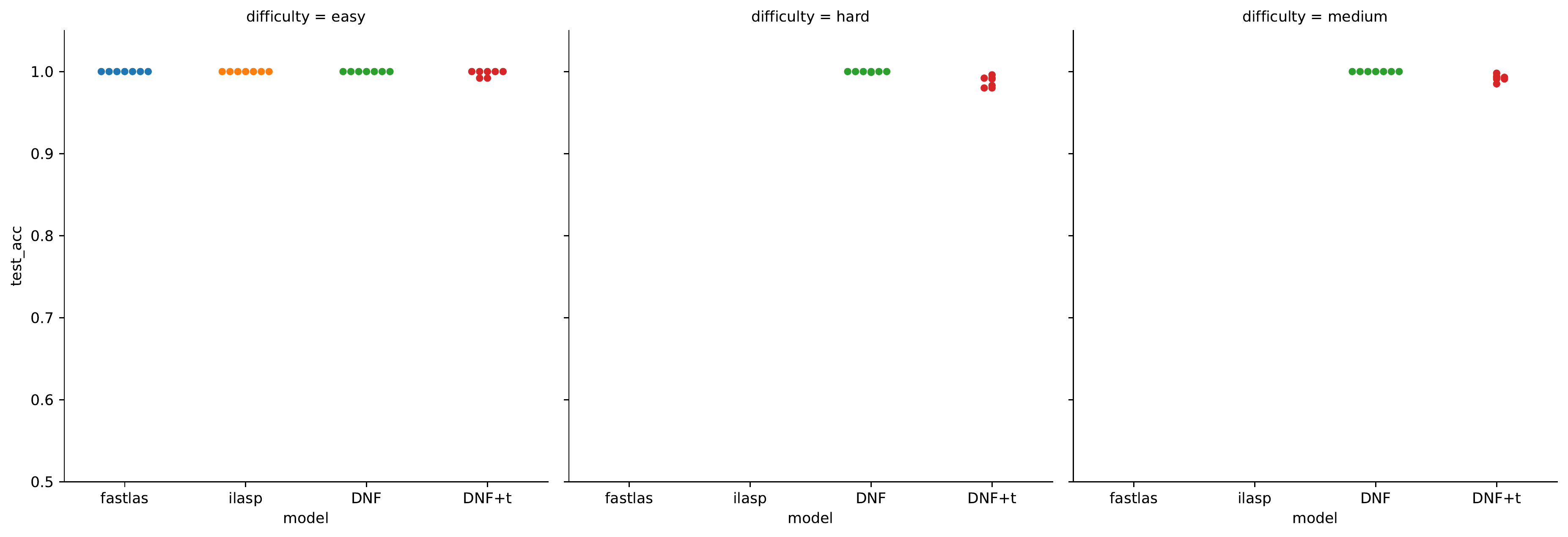}
  \caption{All the data points used in the aggregation of \cref{tab:gendnf_ilp_vs_deep}. There is one point for each random seed used (7).}
  \label{fig:gendnf_ilp_vs_deep}
\end{figure}

\begin{figure}
  \centering
  \renewcommand{\arraystretch}{1.2}
  \begin{tabular}{rl}
    t :- & unary(X,0), not binary(Y,X,0), binary(Y,X,1).                               \\
    t :- & not nullary(0), not unary(X,0), not unary(X,1), not binary(X,Y,0),          \\
         & not binary(X,Y,1), binary(Y,X,0), binary(Y,X,1).                            \\
    t :- & nullary(0), not nullary(1), not unary(X,0), not unary(X,1), not unary(Y,0), \\
         & unary(Y,1), binary(X,Y,0), not binary(Y,X,0), not binary(Y,X,1).
  \end{tabular}
  \caption{Sample rules learnt on the easy set for subgraph set isomorphism dataset. These are thresholded results, i.e. DNF+t model. Random seed used is 3. These rules can be passed onto clingo to solve the unseen examples.}
  \label{fig:example_rules_learnt_gendnf_deep}
\end{figure}

\begin{figure}
  \centering
  \includegraphics[width=1.0\textwidth]{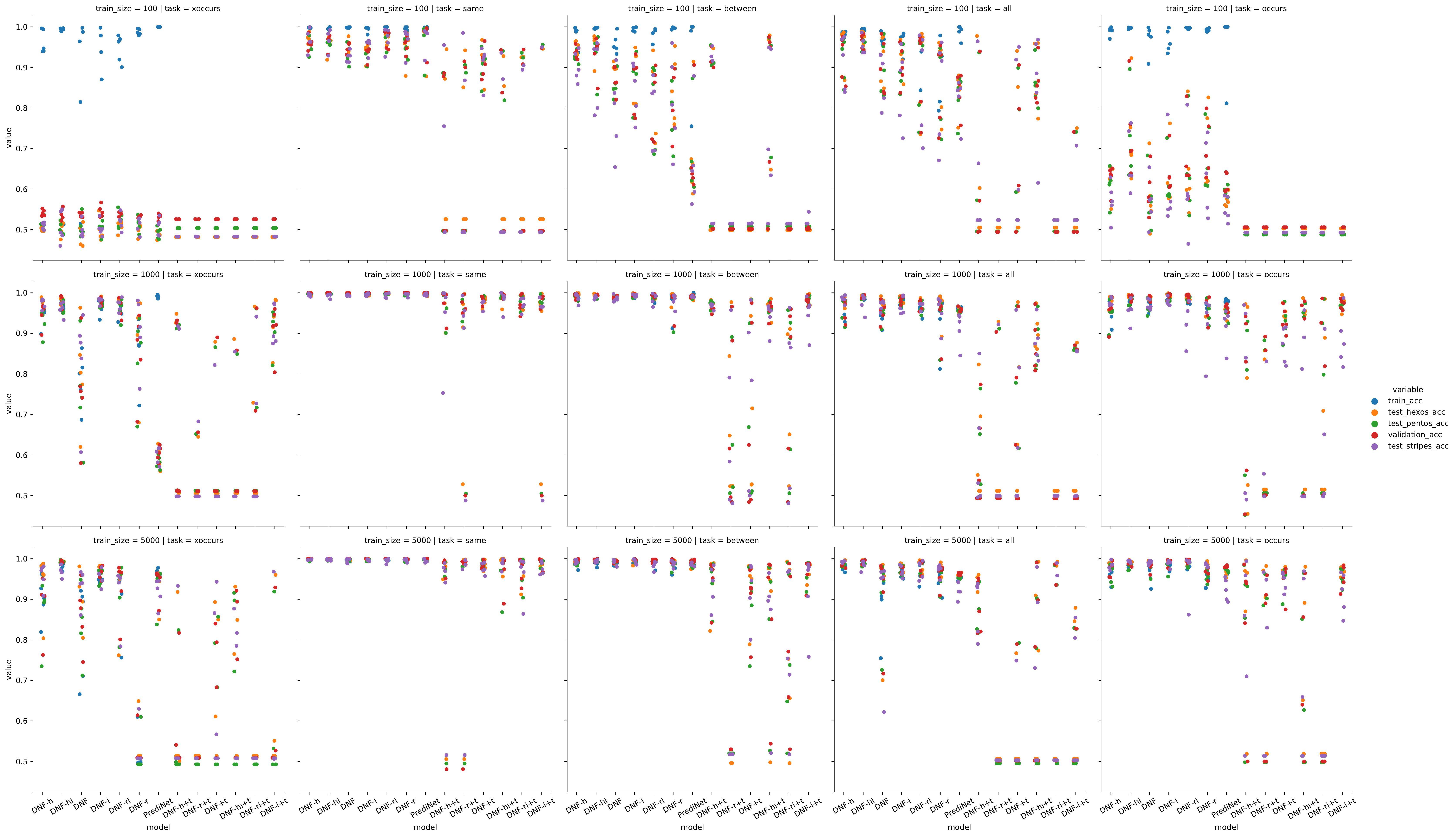}
  \caption{All of the runs on the relations game dataset results, there are 75 PrediNet, 225 DNF and 225 DNF with image reconstruction configurations trained in total. For each DNF configuration there is also the thresholded results noted with suffix \texttt{+t} in the model name.}
  \label{fig:relsgame_all_runs}
\end{figure}

\begin{table}
  \centering
  \caption{Median of test accuracy for all DNF configurations across all relations game setups with mean absolute deviation. Here we observe that the hidden layer version DNF-h has a slight advantage.}
  \label{tab:relsgame_dnf_comparison}
  \begin{tabular}{lrrr}
    \toprule
    Model & Hexominoes               & Pentominoes              & Stripes                  \\
    \midrule
    DNF   & 0.953$\pm$0.186          & 0.932$\pm$0.183          & 0.926$\pm$0.182          \\
    DNF-h & \textbf{0.964}$\pm$0.176 & \textbf{0.943}$\pm$0.172 & \textbf{0.949}$\pm$0.177 \\
    DNF-r & 0.868$\pm$0.214          & 0.863$\pm$0.212          & 0.819$\pm$0.212          \\
    \bottomrule
  \end{tabular}
\end{table}

\begin{table}
  \centering
  \caption{Median of test accuracy for all models across all relations game setups from \cref{tab:relsgame_results}.}
  \label{tab:relsgame_full_results}
  \tiny
  \begin{tabular}{llrrrrrrrrrrrrrrr}
    \toprule
            & task        & \multicolumn{3}{l}{all} & \multicolumn{3}{l}{between} & \multicolumn{3}{l}{occurs} & \multicolumn{3}{l}{same} & \multicolumn{3}{l}{xoccurs}                                                                       \\
            & train\_size & 100                     & 1000                        & 5000                       & 100                      & 1000                        & 5000 & 100  & 1000 & 5000 & 100  & 1000 & 5000 & 100  & 1000 & 5000 \\
    {}      & model       &                         &                             &                            &                          &                             &      &      &      &      &      &      &      &      &      &      \\
    \midrule
    Hex.    & DNF         & 0.94                    & 0.97                        & 0.98                       & 0.90                     & 0.99                        & 0.99 & 0.56 & 0.99 & 0.99 & 0.94 & 1.00 & 1.00 & 0.49 & 0.80 & 0.93 \\
            & DNF+t       & 0.60                    & 0.63                        & 0.51                       & 0.50                     & 0.71                        & 0.96 & 0.50 & 0.94 & 0.97 & 0.92 & 0.98 & 0.99 & 0.48 & 0.51 & 0.61 \\
            & DNF-h       & 0.98                    & 0.99                        & 0.99                       & 0.95                     & 1.00                        & 0.99 & 0.62 & 0.99 & 0.99 & 0.97 & 1.00 & 1.00 & 0.50 & 0.98 & 0.96 \\
            & DNF-h+t     & 0.51                    & 0.55                        & 0.92                       & 0.91                     & 0.97                        & 0.98 & 0.50 & 0.79 & 0.96 & 0.53 & 1.00 & 0.98 & 0.48 & 0.51 & 0.51 \\
            & DNF-hi      & 0.98                    & 0.99                        & 1.00                       & 0.97                     & 0.99                        & 1.00 & 0.69 & 0.99 & 0.99 & 0.97 & 1.00 & 1.00 & 0.51 & 0.99 & 0.99 \\
            & DNF-hi+t    & 0.86                    & 0.90                        & 0.77                       & 0.95                     & 0.97                        & 0.95 & 0.50 & 0.95 & 0.65 & 0.53 & 0.99 & 1.00 & 0.48 & 0.51 & 0.76 \\
            & DNF-i       & 0.93                    & 0.99                        & 0.98                       & 0.93                     & 1.00                        & 1.00 & 0.58 & 0.99 & 0.99 & 0.94 & 1.00 & 1.00 & 0.49 & 0.99 & 0.98 \\
            & DNF-i+t     & 0.51                    & 0.51                        & 0.51                       & 0.50                     & 0.99                        & 0.99 & 0.50 & 0.97 & 0.97 & 0.53 & 0.97 & 0.99 & 0.48 & 0.94 & 0.51 \\
            & DNF-r       & 0.94                    & 0.98                        & 0.98                       & 0.84                     & 1.00                        & 0.99 & 0.63 & 0.99 & 0.99 & 0.96 & 1.00 & 1.00 & 0.51 & 0.94 & 0.51 \\
            & DNF-r+t     & 0.51                    & 0.51                        & 0.51                       & 0.50                     & 0.65                        & 0.50 & 0.50 & 0.51 & 0.90 & 0.53 & 0.95 & 0.99 & 0.48 & 0.51 & 0.51 \\
            & DNF-ri      & 0.96                    & 0.98                        & 0.99                       & 0.88                     & 0.99                        & 1.00 & 0.60 & 0.99 & 0.99 & 0.98 & 1.00 & 1.00 & 0.52 & 0.98 & 0.96 \\
            & DNF-ri+t    & 0.51                    & 0.51                        & 0.51                       & 0.50                     & 0.90                        & 0.75 & 0.50 & 0.71 & 0.52 & 0.53 & 0.97 & 0.99 & 0.48 & 0.51 & 0.51 \\
            & PrediNet    & 0.85                    & 0.95                        & 0.96                       & 0.66                     & 0.99                        & 0.99 & 0.57 & 0.95 & 0.97 & 0.99 & 1.00 & 1.00 & 0.50 & 0.58 & 0.95 \\
    Pent.   & DNF         & 0.89                    & 0.96                        & 0.95                       & 0.85                     & 0.99                        & 0.99 & 0.57 & 0.96 & 0.98 & 0.95 & 1.00 & 1.00 & 0.50 & 0.74 & 0.86 \\
            & DNF+t       & 0.59                    & 0.62                        & 0.50                       & 0.51                     & 0.67                        & 0.92 & 0.49 & 0.93 & 0.96 & 0.91 & 0.98 & 0.99 & 0.50 & 0.51 & 0.68 \\
            & DNF-h       & 0.95                    & 0.97                        & 0.98                       & 0.92                     & 0.99                        & 0.99 & 0.62 & 0.95 & 0.97 & 0.94 & 1.00 & 1.00 & 0.51 & 0.94 & 0.90 \\
            & DNF-h+t     & 0.50                    & 0.53                        & 0.88                       & 0.90                     & 0.98                        & 0.97 & 0.49 & 0.81 & 0.93 & 0.50 & 0.99 & 0.97 & 0.50 & 0.51 & 0.49 \\
            & DNF-hi      & 0.96                    & 0.99                        & 0.99                       & 0.95                     & 0.99                        & 1.00 & 0.69 & 0.98 & 0.98 & 0.96 & 1.00 & 1.00 & 0.50 & 0.96 & 0.99 \\
            & DNF-hi+t    & 0.86                    & 0.85                        & 0.78                       & 0.95                     & 0.96                        & 0.94 & 0.49 & 0.94 & 0.63 & 0.50 & 0.99 & 0.99 & 0.50 & 0.51 & 0.72 \\
            & DNF-i       & 0.89                    & 0.99                        & 0.98                       & 0.88                     & 0.99                        & 0.99 & 0.59 & 0.98 & 0.99 & 0.93 & 1.00 & 1.00 & 0.51 & 0.97 & 0.96 \\
            & DNF-i+t     & 0.50                    & 0.50                        & 0.50                       & 0.51                     & 0.99                        & 0.99 & 0.49 & 0.97 & 0.95 & 0.50 & 0.98 & 0.99 & 0.50 & 0.93 & 0.49 \\
            & DNF-r       & 0.93                    & 0.96                        & 0.97                       & 0.81                     & 0.99                        & 0.99 & 0.65 & 0.96 & 0.96 & 0.95 & 1.00 & 1.00 & 0.52 & 0.88 & 0.49 \\
            & DNF-r+t     & 0.50                    & 0.50                        & 0.50                       & 0.51                     & 0.62                        & 0.52 & 0.49 & 0.51 & 0.88 & 0.50 & 0.96 & 0.98 & 0.50 & 0.51 & 0.49 \\
            & DNF-ri      & 0.95                    & 0.96                        & 0.99                       & 0.86                     & 0.99                        & 1.00 & 0.63 & 0.98 & 0.99 & 0.97 & 1.00 & 1.00 & 0.52 & 0.97 & 0.96 \\
            & DNF-ri+t    & 0.50                    & 0.50                        & 0.50                       & 0.51                     & 0.89                        & 0.74 & 0.49 & 0.80 & 0.50 & 0.50 & 0.98 & 0.99 & 0.50 & 0.51 & 0.49 \\
            & PrediNet    & 0.85                    & 0.96                        & 0.95                       & 0.65                     & 0.99                        & 0.98 & 0.60 & 0.95 & 0.97 & 0.99 & 1.00 & 1.00 & 0.50 & 0.58 & 0.95 \\
    Stripes & DNF         & 0.91                    & 0.97                        & 0.95                       & 0.81                     & 0.98                        & 0.99 & 0.57 & 0.97 & 0.99 & 0.93 & 0.99 & 1.00 & 0.49 & 0.88 & 0.94 \\
            & DNF+t       & 0.60                    & 0.62                        & 0.50                       & 0.51                     & 0.78                        & 0.95 & 0.49 & 0.88 & 0.95 & 0.93 & 0.98 & 0.97 & 0.48 & 0.50 & 0.57 \\
            & DNF-h       & 0.93                    & 0.98                        & 0.99                       & 0.89                     & 0.99                        & 0.99 & 0.57 & 0.97 & 0.99 & 0.96 & 1.00 & 1.00 & 0.51 & 0.98 & 0.97 \\
            & DNF-h+t     & 0.52                    & 0.53                        & 0.93                       & 0.92                     & 0.97                        & 0.95 & 0.49 & 0.84 & 0.86 & 0.49 & 0.99 & 0.97 & 0.48 & 0.50 & 0.51 \\
            & DNF-hi      & 0.95                    & 0.99                        & 0.99                       & 0.94                     & 0.99                        & 0.99 & 0.63 & 0.96 & 0.99 & 0.97 & 1.00 & 1.00 & 0.49 & 0.96 & 0.97 \\
            & DNF-hi+t    & 0.89                    & 0.87                        & 0.73                       & 0.70                     & 0.96                        & 0.90 & 0.49 & 0.81 & 0.66 & 0.49 & 0.99 & 0.99 & 0.48 & 0.50 & 0.79 \\
            & DNF-i       & 0.89                    & 0.96                        & 0.97                       & 0.87                     & 0.99                        & 0.99 & 0.55 & 0.98 & 0.99 & 0.96 & 1.00 & 1.00 & 0.50 & 0.98 & 0.96 \\
            & DNF-i+t     & 0.52                    & 0.50                        & 0.50                       & 0.51                     & 0.96                        & 0.97 & 0.49 & 0.87 & 0.93 & 0.49 & 0.97 & 0.97 & 0.48 & 0.89 & 0.51 \\
            & DNF-r       & 0.92                    & 0.97                        & 0.98                       & 0.81                     & 0.99                        & 0.99 & 0.64 & 0.95 & 0.98 & 0.98 & 1.00 & 1.00 & 0.52 & 0.92 & 0.51 \\
            & DNF-r+t     & 0.52                    & 0.50                        & 0.50                       & 0.51                     & 0.58                        & 0.52 & 0.49 & 0.50 & 0.83 & 0.49 & 0.94 & 0.99 & 0.48 & 0.50 & 0.51 \\
            & DNF-ri      & 0.94                    & 0.97                        & 0.99                       & 0.84                     & 0.99                        & 0.99 & 0.58 & 0.95 & 0.99 & 0.98 & 1.00 & 1.00 & 0.52 & 0.98 & 0.95 \\
            & DNF-ri+t    & 0.52                    & 0.50                        & 0.50                       & 0.51                     & 0.87                        & 0.75 & 0.49 & 0.50 & 0.51 & 0.49 & 0.97 & 0.98 & 0.48 & 0.50 & 0.51 \\
            & PrediNet    & 0.84                    & 0.93                        & 0.92                       & 0.64                     & 0.99                        & 0.99 & 0.54 & 0.94 & 0.92 & 0.99 & 0.99 & 1.00 & 0.51 & 0.61 & 0.93 \\
    \bottomrule
  \end{tabular}
\end{table}

\begin{table}
  \centering
  \caption{Median absolute deviation of test accuracy for all models across all relations game setups, complements \cref{tab:relsgame_full_results}.}
  \label{tab:relations_game_full_results_mad}
  \tiny
  \begin{tabular}{llrrrrrrrrrrrrrrr}
    \toprule
            & task        & \multicolumn{3}{l}{all} & \multicolumn{3}{l}{between} & \multicolumn{3}{l}{occurs} & \multicolumn{3}{l}{same} & \multicolumn{3}{l}{xoccurs}                                                                       \\
            & train\_size & 100                     & 1000                        & 5000                       & 100                      & 1000                        & 5000 & 100  & 1000 & 5000 & 100  & 1000 & 5000 & 100  & 1000 & 5000 \\
    {}      & model       &                         &                             &                            &                          &                             &      &      &      &      &      &      &      &      &      &      \\
    \midrule
    Hex.    & DNF         & 0.05                    & 0.01                        & 0.09                       & 0.03                     & 0.00                        & 0.00 & 0.06 & 0.00 & 0.00 & 0.01 & 0.00 & 0.00 & 0.02 & 0.08 & 0.05 \\
            & DNF+t       & 0.17                    & 0.17                        & 0.08                       & 0.00                     & 0.18                        & 0.06 & 0.00 & 0.01 & 0.00 & 0.03 & 0.01 & 0.01 & 0.00 & 0.12 & 0.16 \\
            & DNF-h       & 0.05                    & 0.01                        & 0.00                       & 0.01                     & 0.00                        & 0.00 & 0.03 & 0.01 & 0.00 & 0.01 & 0.00 & 0.00 & 0.00 & 0.01 & 0.05 \\
            & DNF-h+t     & 0.14                    & 0.11                        & 0.06                       & 0.21                     & 0.01                        & 0.05 & 0.00 & 0.20 & 0.14 & 0.18 & 0.03 & 0.15 & 0.00 & 0.21 & 0.13 \\
            & DNF-hi      & 0.01                    & 0.00                        & 0.00                       & 0.03                     & 0.00                        & 0.00 & 0.08 & 0.00 & 0.00 & 0.02 & 0.00 & 0.00 & 0.01 & 0.01 & 0.00 \\
            & DNF-hi+t    & 0.07                    & 0.04                        & 0.18                       & 0.19                     & 0.01                        & 0.15 & 0.00 & 0.21 & 0.18 & 0.18 & 0.01 & 0.02 & 0.00 & 0.12 & 0.16 \\
            & DNF-i       & 0.03                    & 0.01                        & 0.01                       & 0.06                     & 0.00                        & 0.00 & 0.06 & 0.00 & 0.00 & 0.01 & 0.00 & 0.00 & 0.02 & 0.00 & 0.01 \\
            & DNF-i+t     & 0.08                    & 0.17                        & 0.17                       & 0.00                     & 0.01                        & 0.02 & 0.00 & 0.01 & 0.01 & 0.13 & 0.14 & 0.01 & 0.00 & 0.05 & 0.14 \\
            & DNF-r       & 0.08                    & 0.03                        & 0.01                       & 0.07                     & 0.01                        & 0.00 & 0.09 & 0.02 & 0.01 & 0.03 & 0.00 & 0.00 & 0.01 & 0.08 & 0.04 \\
            & DNF-r+t     & 0.00                    & 0.13                        & 0.00                       & 0.00                     & 0.17                        & 0.00 & 0.00 & 0.16 & 0.21 & 0.18 & 0.14 & 0.16 & 0.00 & 0.04 & 0.00 \\
            & DNF-ri      & 0.10                    & 0.00                        & 0.01                       & 0.09                     & 0.00                        & 0.00 & 0.08 & 0.00 & 0.00 & 0.01 & 0.00 & 0.00 & 0.01 & 0.00 & 0.06 \\
            & DNF-ri+t    & 0.00                    & 0.00                        & 0.23                       & 0.00                     & 0.16                        & 0.17 & 0.00 & 0.17 & 0.00 & 0.19 & 0.01 & 0.03 & 0.00 & 0.16 & 0.00 \\
            & PrediNet    & 0.04                    & 0.00                        & 0.00                       & 0.09                     & 0.00                        & 0.01 & 0.01 & 0.01 & 0.01 & 0.04 & 0.00 & 0.00 & 0.01 & 0.02 & 0.03 \\
    Pent.   & DNF         & 0.05                    & 0.02                        & 0.07                       & 0.02                     & 0.00                        & 0.00 & 0.05 & 0.02 & 0.01 & 0.02 & 0.00 & 0.00 & 0.02 & 0.08 & 0.06 \\
            & DNF+t       & 0.15                    & 0.16                        & 0.10                       & 0.00                     & 0.19                        & 0.07 & 0.00 & 0.03 & 0.02 & 0.03 & 0.01 & 0.01 & 0.00 & 0.11 & 0.14 \\
            & DNF-h       & 0.05                    & 0.02                        & 0.00                       & 0.01                     & 0.00                        & 0.00 & 0.03 & 0.02 & 0.02 & 0.01 & 0.00 & 0.00 & 0.01 & 0.02 & 0.06 \\
            & DNF-h+t     & 0.13                    & 0.10                        & 0.05                       & 0.20                     & 0.01                        & 0.04 & 0.00 & 0.18 & 0.14 & 0.19 & 0.03 & 0.15 & 0.00 & 0.19 & 0.11 \\
            & DNF-hi      & 0.01                    & 0.00                        & 0.00                       & 0.04                     & 0.00                        & 0.00 & 0.07 & 0.01 & 0.01 & 0.01 & 0.00 & 0.00 & 0.01 & 0.01 & 0.00 \\
            & DNF-hi+t    & 0.04                    & 0.05                        & 0.19                       & 0.18                     & 0.01                        & 0.13 & 0.00 & 0.22 & 0.18 & 0.18 & 0.00 & 0.04 & 0.00 & 0.11 & 0.17 \\
            & DNF-i       & 0.04                    & 0.01                        & 0.01                       & 0.06                     & 0.00                        & 0.00 & 0.04 & 0.00 & 0.01 & 0.01 & 0.00 & 0.00 & 0.01 & 0.01 & 0.02 \\
            & DNF-i+t     & 0.08                    & 0.18                        & 0.16                       & 0.00                     & 0.01                        & 0.03 & 0.00 & 0.01 & 0.02 & 0.15 & 0.15 & 0.01 & 0.00 & 0.04 & 0.13 \\
            & DNF-r       & 0.09                    & 0.04                        & 0.02                       & 0.07                     & 0.03                        & 0.01 & 0.07 & 0.02 & 0.01 & 0.02 & 0.00 & 0.00 & 0.01 & 0.08 & 0.04 \\
            & DNF-r+t     & 0.00                    & 0.13                        & 0.00                       & 0.00                     & 0.18                        & 0.00 & 0.00 & 0.18 & 0.20 & 0.19 & 0.15 & 0.16 & 0.00 & 0.04 & 0.00 \\
            & DNF-ri      & 0.09                    & 0.01                        & 0.01                       & 0.09                     & 0.00                        & 0.00 & 0.08 & 0.00 & 0.01 & 0.02 & 0.00 & 0.00 & 0.02 & 0.02 & 0.06 \\
            & DNF-ri+t    & 0.00                    & 0.00                        & 0.22                       & 0.00                     & 0.17                        & 0.16 & 0.00 & 0.19 & 0.00 & 0.21 & 0.01 & 0.03 & 0.00 & 0.16 & 0.00 \\
            & PrediNet    & 0.04                    & 0.00                        & 0.00                       & 0.08                     & 0.00                        & 0.01 & 0.01 & 0.01 & 0.00 & 0.04 & 0.00 & 0.00 & 0.01 & 0.01 & 0.04 \\
    Stripes & DNF         & 0.06                    & 0.01                        & 0.11                       & 0.08                     & 0.01                        & 0.01 & 0.04 & 0.01 & 0.00 & 0.01 & 0.00 & 0.00 & 0.01 & 0.09 & 0.03 \\
            & DNF+t       & 0.19                    & 0.17                        & 0.08                       & 0.00                     & 0.18                        & 0.06 & 0.00 & 0.04 & 0.02 & 0.03 & 0.01 & 0.01 & 0.00 & 0.10 & 0.18 \\
            & DNF-h       & 0.06                    & 0.01                        & 0.00                       & 0.03                     & 0.01                        & 0.00 & 0.03 & 0.01 & 0.01 & 0.02 & 0.00 & 0.00 & 0.01 & 0.01 & 0.02 \\
            & DNF-h+t     & 0.14                    & 0.12                        & 0.06                       & 0.20                     & 0.00                        & 0.04 & 0.00 & 0.20 & 0.16 & 0.17 & 0.07 & 0.15 & 0.00 & 0.20 & 0.14 \\
            & DNF-hi      & 0.02                    & 0.02                        & 0.01                       & 0.08                     & 0.00                        & 0.00 & 0.06 & 0.02 & 0.00 & 0.02 & 0.00 & 0.00 & 0.03 & 0.01 & 0.01 \\
            & DNF-hi+t    & 0.10                    & 0.03                        & 0.17                       & 0.16                     & 0.02                        & 0.12 & 0.00 & 0.19 & 0.19 & 0.20 & 0.02 & 0.01 & 0.00 & 0.11 & 0.15 \\
            & DNF-i       & 0.08                    & 0.02                        & 0.01                       & 0.06                     & 0.00                        & 0.00 & 0.07 & 0.01 & 0.01 & 0.01 & 0.00 & 0.00 & 0.01 & 0.01 & 0.02 \\
            & DNF-i+t     & 0.06                    & 0.17                        & 0.16                       & 0.01                     & 0.03                        & 0.07 & 0.00 & 0.05 & 0.04 & 0.15 & 0.15 & 0.01 & 0.00 & 0.04 & 0.15 \\
            & DNF-r       & 0.11                    & 0.03                        & 0.01                       & 0.09                     & 0.01                        & 0.01 & 0.08 & 0.05 & 0.00 & 0.02 & 0.00 & 0.00 & 0.02 & 0.06 & 0.04 \\
            & DNF-r+t     & 0.00                    & 0.14                        & 0.00                       & 0.00                     & 0.17                        & 0.00 & 0.00 & 0.10 & 0.19 & 0.21 & 0.15 & 0.15 & 0.00 & 0.06 & 0.00 \\
            & DNF-ri      & 0.11                    & 0.01                        & 0.02                       & 0.08                     & 0.01                        & 0.01 & 0.08 & 0.04 & 0.04 & 0.02 & 0.00 & 0.00 & 0.01 & 0.01 & 0.05 \\
            & DNF-ri+t    & 0.00                    & 0.00                        & 0.23                       & 0.00                     & 0.19                        & 0.15 & 0.00 & 0.13 & 0.00 & 0.19 & 0.01 & 0.04 & 0.00 & 0.16 & 0.00 \\
            & PrediNet    & 0.04                    & 0.03                        & 0.02                       & 0.08                     & 0.00                        & 0.00 & 0.03 & 0.03 & 0.02 & 0.02 & 0.00 & 0.00 & 0.02 & 0.02 & 0.03 \\
    \bottomrule
  \end{tabular}
\end{table}

\begin{table}
  \centering
  \caption{Aggregate median test accuracy for all DNF models with and without image reconstruction loss along with median absolute deviation. Only for XOccurs task do we see an improvement.}
  \label{tab:relsgame_image_reconstruction_results}
  \footnotesize
  \begin{tabular}{llrrrrrrrrrrrr}
    \toprule
                                             & {}    & \multicolumn{2}{l}{Hexos} & \multicolumn{2}{l}{Pentos} & \multicolumn{2}{l}{Stripes}                                                 \\
    \multicolumn{2}{r}{Image Reconsturction} & False & True                      & False                      & True                        & False         & True                          \\
    \midrule
    all                                      & 100   & 0.94$\pm$0.06             & 0.96$\pm$0.05              & 0.93$\pm$0.06               & 0.94$\pm$0.06 & 0.92$\pm$0.07 & 0.94$\pm$0.08 \\
                                             & 1000  & 0.98$\pm$0.02             & 0.99$\pm$0.01              & 0.96$\pm$0.03               & 0.98$\pm$0.01 & 0.98$\pm$0.02 & 0.97$\pm$0.01 \\
                                             & 5000  & 0.99$\pm$0.04             & 0.99$\pm$0.01              & 0.97$\pm$0.04               & 0.98$\pm$0.01 & 0.98$\pm$0.04 & 0.98$\pm$0.01 \\
    between                                  & 100   & 0.91$\pm$0.05             & 0.93$\pm$0.07              & 0.87$\pm$0.05               & 0.89$\pm$0.07 & 0.86$\pm$0.08 & 0.84$\pm$0.07 \\
                                             & 1000  & 0.99$\pm$0.01             & 1.00$\pm$0.00              & 0.99$\pm$0.01               & 0.99$\pm$0.00 & 0.99$\pm$0.01 & 0.99$\pm$0.01 \\
                                             & 5000  & 0.99$\pm$0.00             & 1.00$\pm$0.00              & 0.99$\pm$0.00               & 0.99$\pm$0.00 & 0.99$\pm$0.01 & 0.99$\pm$0.01 \\
    occurs                                   & 100   & 0.62$\pm$0.06             & 0.63$\pm$0.09              & 0.61$\pm$0.06               & 0.63$\pm$0.08 & 0.57$\pm$0.06 & 0.59$\pm$0.08 \\
                                             & 1000  & 0.99$\pm$0.01             & 0.99$\pm$0.00              & 0.96$\pm$0.02               & 0.98$\pm$0.01 & 0.97$\pm$0.03 & 0.97$\pm$0.02 \\
                                             & 5000  & 0.99$\pm$0.01             & 0.99$\pm$0.00              & 0.97$\pm$0.02               & 0.98$\pm$0.01 & 0.99$\pm$0.01 & 0.99$\pm$0.02 \\
    same                                     & 100   & 0.96$\pm$0.02             & 0.96$\pm$0.02              & 0.94$\pm$0.02               & 0.94$\pm$0.02 & 0.96$\pm$0.03 & 0.96$\pm$0.02 \\
                                             & 1000  & 1.00$\pm$0.00             & 1.00$\pm$0.00              & 1.00$\pm$0.00               & 1.00$\pm$0.00 & 1.00$\pm$0.00 & 1.00$\pm$0.00 \\
                                             & 5000  & 1.00$\pm$0.00             & 1.00$\pm$0.00              & 1.00$\pm$0.00               & 1.00$\pm$0.00 & 1.00$\pm$0.00 & 1.00$\pm$0.00 \\
    xoccurs                                  & 100   & 0.50$\pm$0.01             & 0.51$\pm$0.02              & 0.51$\pm$0.01               & 0.51$\pm$0.02 & 0.50$\pm$0.01 & 0.50$\pm$0.02 \\
                                             & 1000  & 0.96$\pm$0.10             & 0.98$\pm$0.00              & 0.88$\pm$0.10               & 0.97$\pm$0.01 & 0.94$\pm$0.07 & 0.98$\pm$0.01 \\
                                             & 5000  & 0.89$\pm$0.17             & 0.98$\pm$0.03              & 0.82$\pm$0.16               & 0.97$\pm$0.03 & 0.94$\pm$0.18 & 0.96$\pm$0.03 \\
    \bottomrule
  \end{tabular}
\end{table}

\begin{figure}
  \centering
  \includegraphics[width=0.9\textwidth]{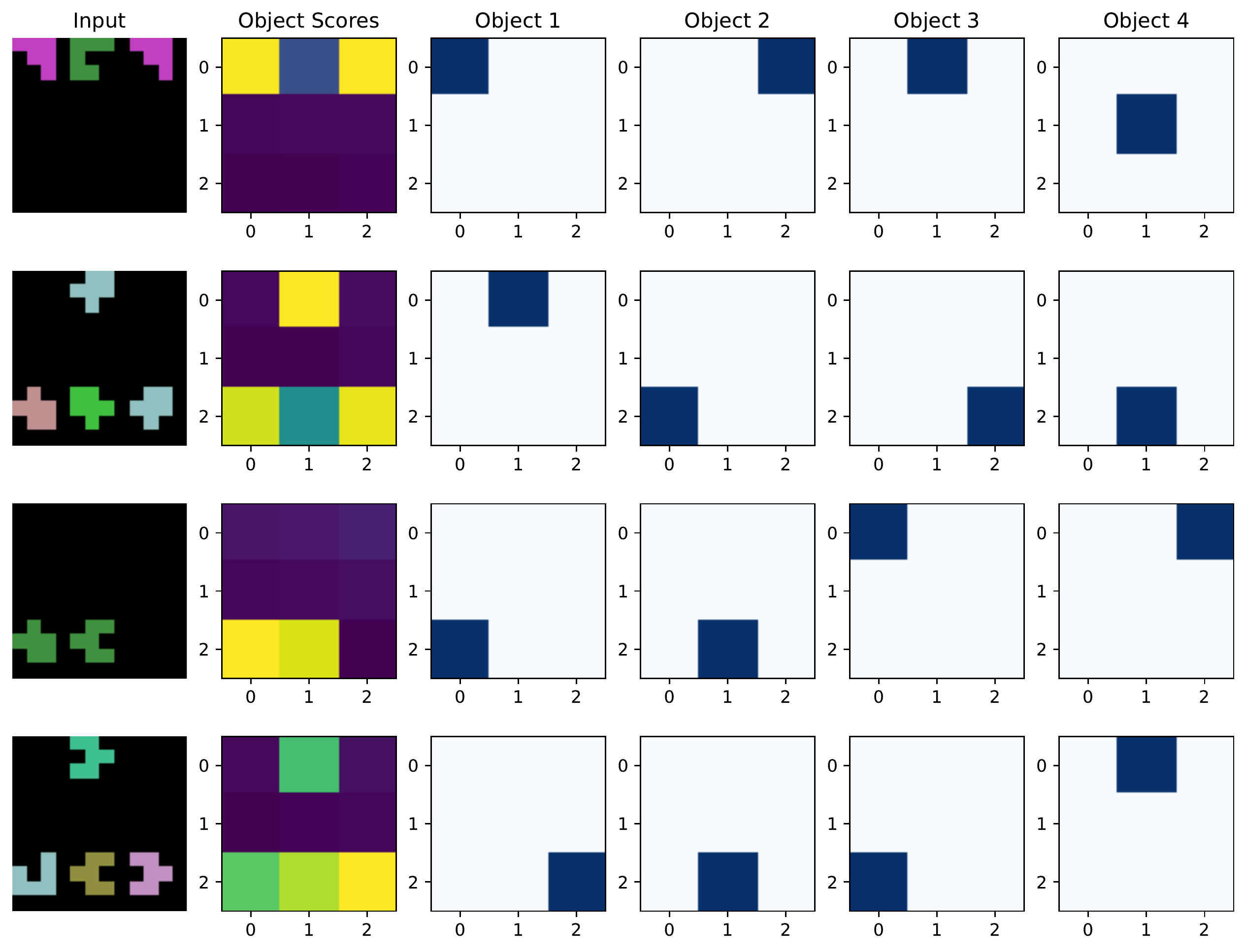}
  \caption{Attention maps with the DNF model trained on all the tasks with 1k training size per task. When there are extra slots, the model is forced to select an empty black patch as an object.}
  \label{fig:relsgame_all_att_maps}
\end{figure}

\begin{table}
  \centering
  \caption{Example pruning steps of the disjunctive SL weights for DNF model with image reconstruction trained on 1k examples on the relations game dataset. This is the last set of weights that predict the output. We report the test pentominoes accuracy after each step. Thresholding the weights often causes the biggest drop as a small weight can be amplified increasing any errors. The single weight 6.0 in the last row correspond to the rule shown in \cref{sec:analysis} for the between task. The accuracy might vary due to stochastic object selection despite having fixed weights, e.g. in the last two rows.}
  \label{tab:relsgame_DNF-i_or_kernel_pruning}
  \footnotesize
  \begin{tabular}{rcl}
    \toprule
    Stage              & Weights                & Test Pent. Acc \\
    \midrule
    Preprune           & \verb|[ 1.56 -1.72  7.45 -2.22 -2.27  2.76 -1.46  1.75]| & 0.996          \\
    Pruned             & \verb|[ 1.56 -1.72  7.45 -0.00  0.00  0.00 -1.46  1.75]| & 0.991          \\
    Threshold          & \verb|[ 0.00 -0.00  6.00  0.00  0.00  0.00 -0.00  0.00]| & 0.942          \\
    Threshold + Pruned & \verb|[ 0.00 -0.00  6.00  0.00  0.00  0.00 -0.00  0.00]| & 0.938          \\
    \bottomrule
  \end{tabular}
\end{table}

\begin{figure}
  \centering
  \includegraphics[width=0.9\textwidth]{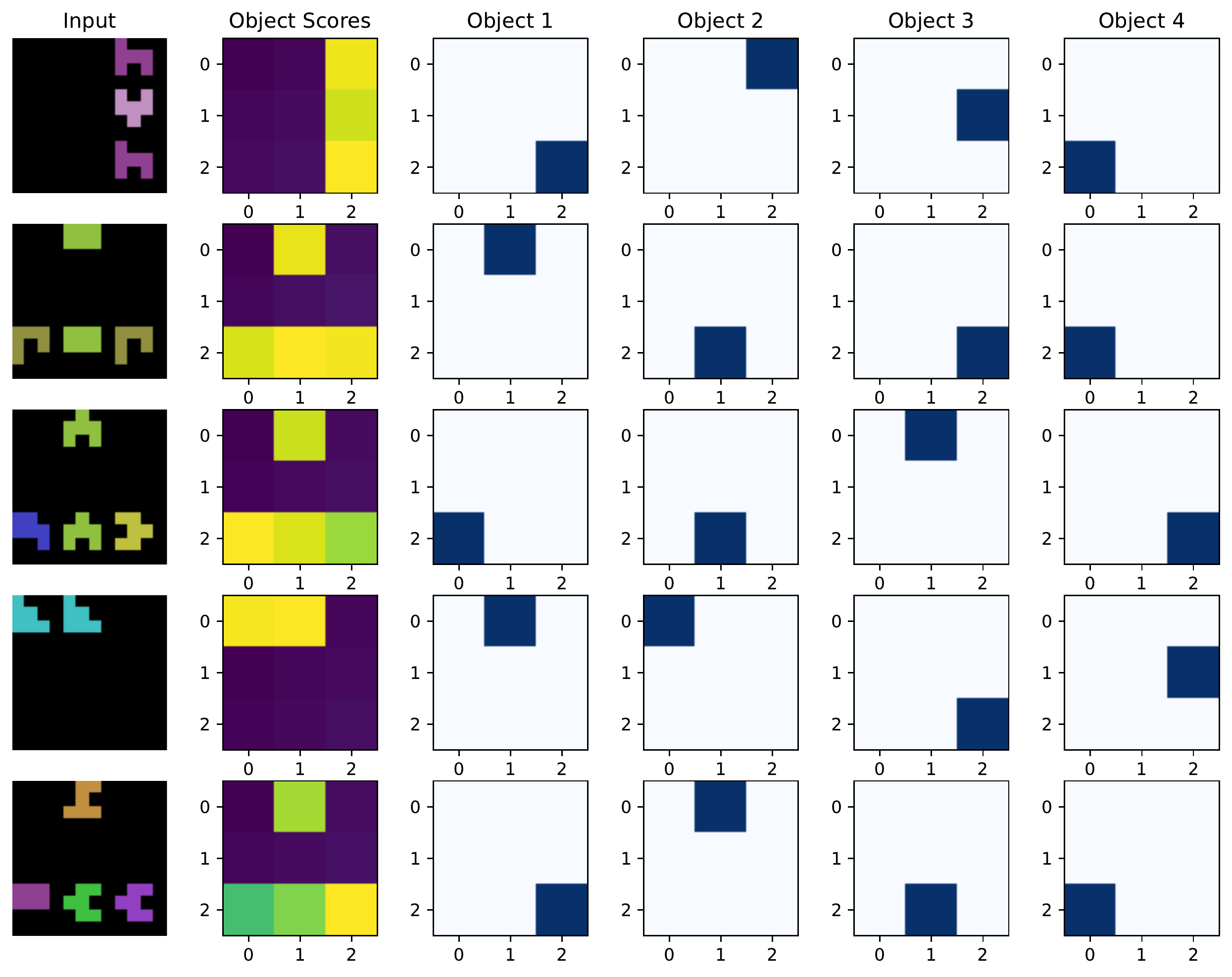}
  \caption{Further attention maps learnt by DNF-hi on all tasks with 1k training size per task.}
  \label{fig:relsgame_DNF-hi_all_att_maps}
\end{figure}

\begin{table}
  \centering
  \caption{Learnt logic program of a successful run with DNF-hi on all tasks with a training size of 1k. The following logic program can be passed to clingo, along with a thresholded interpretation from the neural network to predict the answer. Note, that nullary(0), ..., nullary(3) corresponds to the task ids for Same, Between, Occurs and XOccurs respectively. Since this is a multi-task setting, we observe that the model is utilising the nullary predicates in the logic program. The obj predicate is added during post-processing to ensure all rules are safe.}
  \label{tab:relsgame_DNF-hi_all_learnt_rules}
  \begin{tabular}{|rl|}
    \toprule
    unary(V0,10) :-       & not c3unary(V0,10), obj(V0).                                            \\
    c3unary(V0,10) :-     & not nullary(3), unary(V0,3), binary(V0,V1,6), not binary(V1,V0,1),      \\
                          & binary(V1,V0,3), binary(V1,V0,10), binary(V1,V0,14),                    \\
                          & obj(V1), V1 != V0, obj(V0).                                             \\
    unary(V0,11) :-       & not c1unary(V0,11), obj(V0).                                            \\
    c1unary(V0,11) :-     & not nullary(3), not binary(V0,V1,1), not binary(V0,V1,9),               \\
                          & not binary(V0,V1,11), binary(V1,V0,10), not binary(V1,V0,11),           \\
                          & binary(V1,V0,13), binary(V1,V0,14), obj(V1), V1 != V0, obj(V0).         \\
    binary(V0,V1,23) :-   & not binary(V1,V0,13), obj(V1), V1 != V0, obj(V0).                       \\
    binary(V0,V1,23) :-   & not c3binary(V0,V1,23), obj(V0), obj(V1), V0 != V1.                     \\
    c3binary(V0,V1,23) :- & not nullary(0), not binary(V0,V1,1), binary(V0,V1,3), binary(V0,V1,15), \\
                          & not binary(V1,V0,1), binary(V1,V0,10), not binary(V1,V0,11),            \\
                          & binary(V1,V0,13), binary(V1,V0,14), obj(V1), V1 != V0, obj(V0).         \\
    t :-                  & not c4t.                                                                \\
    c4t :-                & unary(V2,10), unary(V2,11), obj(V2).                                    \\
    t :-                  & binary(V0,V1,23), not binary(V3,V2,23), obj(V2), V2 != V1, V2 != V3,    \\
                          & V2 != V0, obj(V1), V1 != V3, V1 != V0, obj(V3), V3 != V0, obj(V0).      \\
    \bottomrule
  \end{tabular}
\end{table}

\begin{figure}
  \centering
  \includegraphics[width=0.9\textwidth]{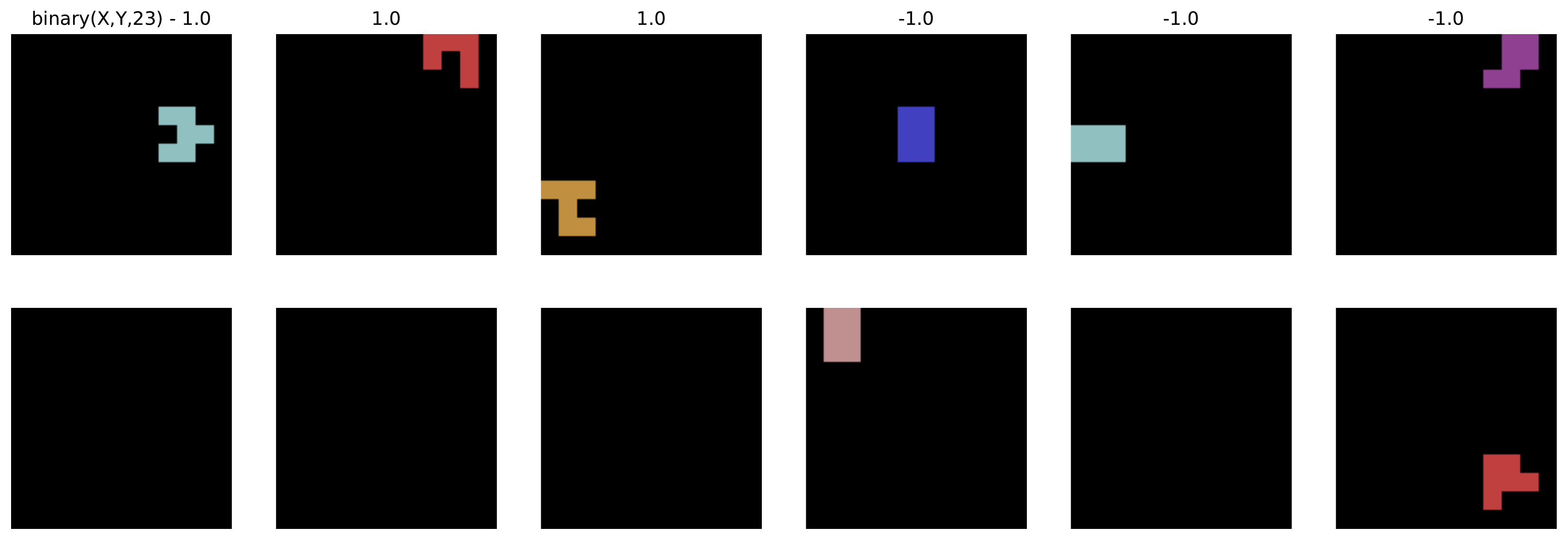}
  \caption{Example truth values for binary(X,Y,23) predicate from the logic program in \cref{tab:relsgame_DNF-hi_all_learnt_rules}. Removing `not binary(V3,V2,23)' from the logic program results in the most accuracy loss, 0.39 .}
  \label{fig:relsgame_binary23_truth_cases}
\end{figure}

\end{document}